\documentclass[times, review, 10 pt]{elsarticle}
\usepackage{amssymb}
\usepackage{lineno}
\usepackage{multirow}
\usepackage{bbm} 
\usepackage{soul}
\usepackage{amsmath}
\usepackage{color}
\usepackage{hyperref}

\usepackage{amsmath}
\usepackage{amssymb}
\usepackage{amsfonts}
\usepackage{booktabs, makecell, multirow, tabularx}
\usepackage{textcomp}
\usepackage{stfloats}
\usepackage{url}
\usepackage{verbatim}
\usepackage{graphicx}
\usepackage{xcolor}
\usepackage{spacingtricks}
\usepackage{bbm}
\usepackage{multirow}
\usepackage{makecell}
\usepackage{graphicx}
\usepackage{amsmath}
\usepackage{booktabs}
\usepackage{multirow}
\usepackage{array}
\usepackage{amsfonts}
\usepackage{bbm} 
\usepackage{xcolor}
\usepackage{mdframed}
\usepackage{makecell}
\usepackage[normalem]{ulem}
\usepackage{graphicx}
\usepackage{subcaption}
\usepackage{amsmath,amsfonts,bm}

\usepackage{setspace}

\usepackage{amsmath,amsfonts,bm}

\usepackage{nicefrac}       
\usepackage{microtype}      
\usepackage{color}
\usepackage{array}
\usepackage{multirow}
\usepackage{dirtytalk}

\usepackage{amsthm}
\usepackage[para,online,flushleft]{threeparttable}

\newcommand{\bb}{{\mathbf{b}}}

\newcommand{\bg}{{\mathbf{g}}}

\newcommand{\bn}{{\mathbf{n}}}

\newcommand{\cL}{{\mathcal{L}}}

\newcommand{\cN}{{\mathcal{N}}}

\newcommand{\btheta}{\bm{\theta}}

\def\argmin{\mathop{\mathrm{arg}\, \mathrm{min}}\limits}

\newcommand\mytab[1]{\begin{tabular}[t]{@{}c@{}} #1 \end{tabular}}
\newcommand\mc[2]{\multicolumn{#1}{c}{#2}}

\newcommand{\nk}{\kern-0.1em}
\journal{Pattern Recognition}

\begin{document}

\begin{frontmatter}

\title{
\textcolor{black}{Ultra-Short rPPG Estimation via Periodicity Guidance and Signal Reconstruction}
\tnoteref{t1}}

\author[1,2]{Pei-Kai Huang\fnref{fn1}}
\ead{alwayswithme@fjnu.edu.cn,alwayswithme@gapp.nthu.edu.tw}

\author[2]{Ya-Ting Chan}
\ead{ytchann@gapp.nthu.edu.tw}

\author[2]{Kuan-Wen Chen}
\ead{cs112062652@gapp.nthu.edu.tw} 

\author[2]{Chiou-Ting Hsu}
\ead{cthsu@cs.nthu.edu.tw}

\author[1]{Xiaoding Wang \corref{cor1}}
\ead{wangdin1982@fjnu.edu.cn} 

\author[3]{Md. Jalil Piran\corref{cor1}}
\ead{piran@sejong.ac.kr}

\cortext[cor1]{Corresponding author}

\affiliation[1]{organization={College of Computer and Cyber Security, Fujian Normal University},
                city={Fuzhou},
                country={China}}

\affiliation[2]{organization={Department of Computer Science, 
                              National Tsing Hua University},
                city={Hsinchu},
                country={Taiwan}}

\affiliation[3]{organization={Department of Computer Science and Engineering, Seojg University}, city={Seoul}, city={South Korea}}
\begin{abstract}
Many remote Heart Rate (HR) measurement methods focus on estimating remote photoplethysmography (rPPG) signals from video clips lasting around 10 seconds but often overlook the need for HR estimation from ultra-short video clips.
In this paper, we aim to accurately measure HR from ultra-short 2-second video clips by specifically addressing two key challenges.
First, to overcome the limited number of heartbeat cycles in ultra-short video clips, we propose an effective periodicity-guided rPPG estimation method that enforces consistent periodicity between rPPG signals estimated from ultra-short clips and their much longer ground truth signals.
Next, to mitigate estimation inaccuracies due to spectral leakage, we propose including a generator to reconstruct longer rPPG signals from ultra-short ones while preserving their periodic consistency to enable more accurate HR measurement.
Extensive experiments on four rPPG estimation benchmark datasets demonstrate that our proposed method not only accurately measures HR from ultra-short video clips but also \textcolor{black}{outperforms} previous rPPG estimation techniques to achieve state-of-the-art performance.
\end{abstract}

\end{frontmatter}


\section{Introduction}
\label{sec:intro}

Heart rate (HR) measurement using remote photoplethysmography (rPPG) \cite{shao2023tranphys,huang2021spatio,liu2024rppg} is a non-contact technique that analyzes subtle changes in skin color caused by blood flow. Recent advancements in remote HR measurement methods \cite{zhang2024self, sun2022contrast, speth2023non} primarily focus on estimating rPPG signals from video clips lasting around 10 seconds. 
Since the human heart rate typically ranges from 40 to 250 beats per minute (bpm), the power spectral densities (PSDs) of the estimated rPPG signals are expected to fall within the frequency range of 0.66 Hz to 4.16 Hz \cite{gideon2021way, sun2022contrast}.
Therefore, it is practicable to identify the frequency corresponding to the highest peak in the PSDs of the estimated rPPG signals from 10-second video clips to determine the HR value \cite{gideon2021way, sun2022contrast}. 
\textcolor{black}{However, estimating rPPG signals from 10-second video clips may fail to provide timely information about sudden HR changes, such as those associated with fatigue driving \cite{9785381, 9036934} or other acute physiological events. In contrast, HR measurement from ultra-short video clips, such as 2-second clips, would be more effective for delivering immediate information and enabling prompt interventions for early warning \cite{dang2019survey}.}

\begin{figure}[t]  
    \centering
    \includegraphics[width= 8.5 cm]{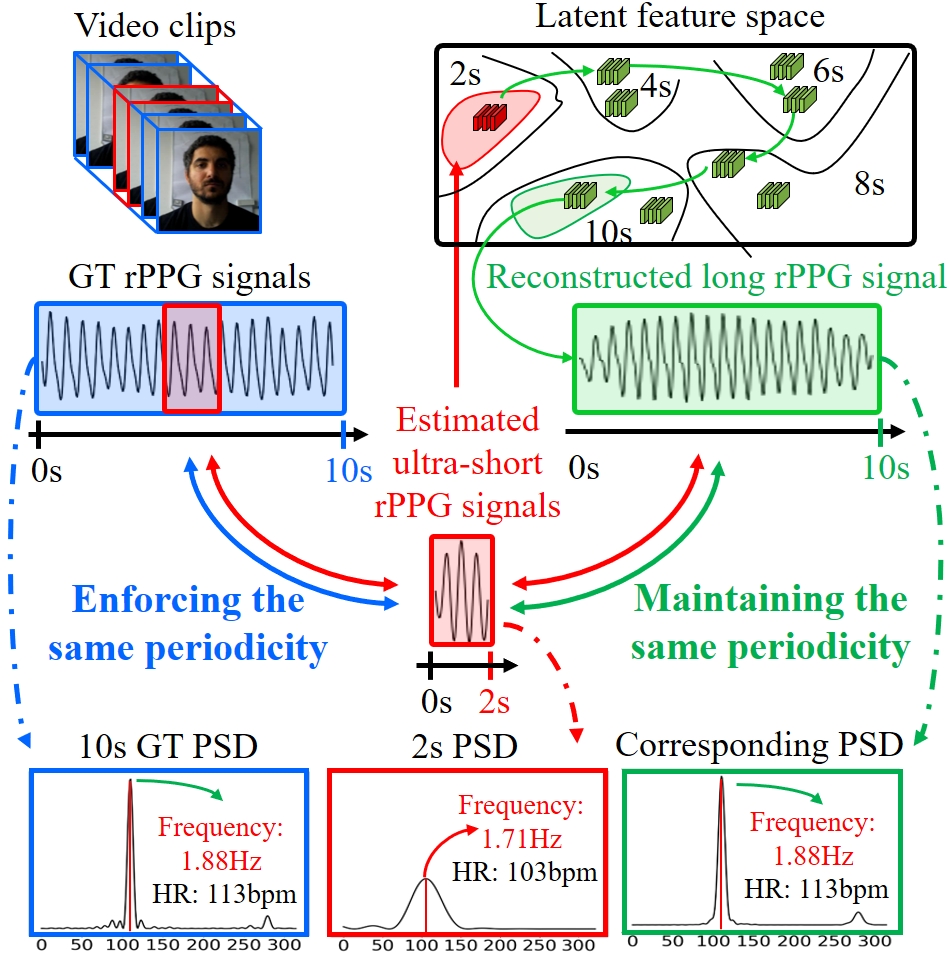} 
\caption{  
The proposed method estimates heart rate from ultra-short 2-second video clips by enforcing periodicity consistency and reconstructing longer rPPG signals, reducing spectral leakage for more accurate PSD estimation and heart rate measurement.
} 
    \label{fig:idea}  
\end{figure} 

\begin{figure}[t]  
    \centering
    \begin{tabular}{cc} 
    \fbox{\includegraphics[width=0.375\columnwidth]{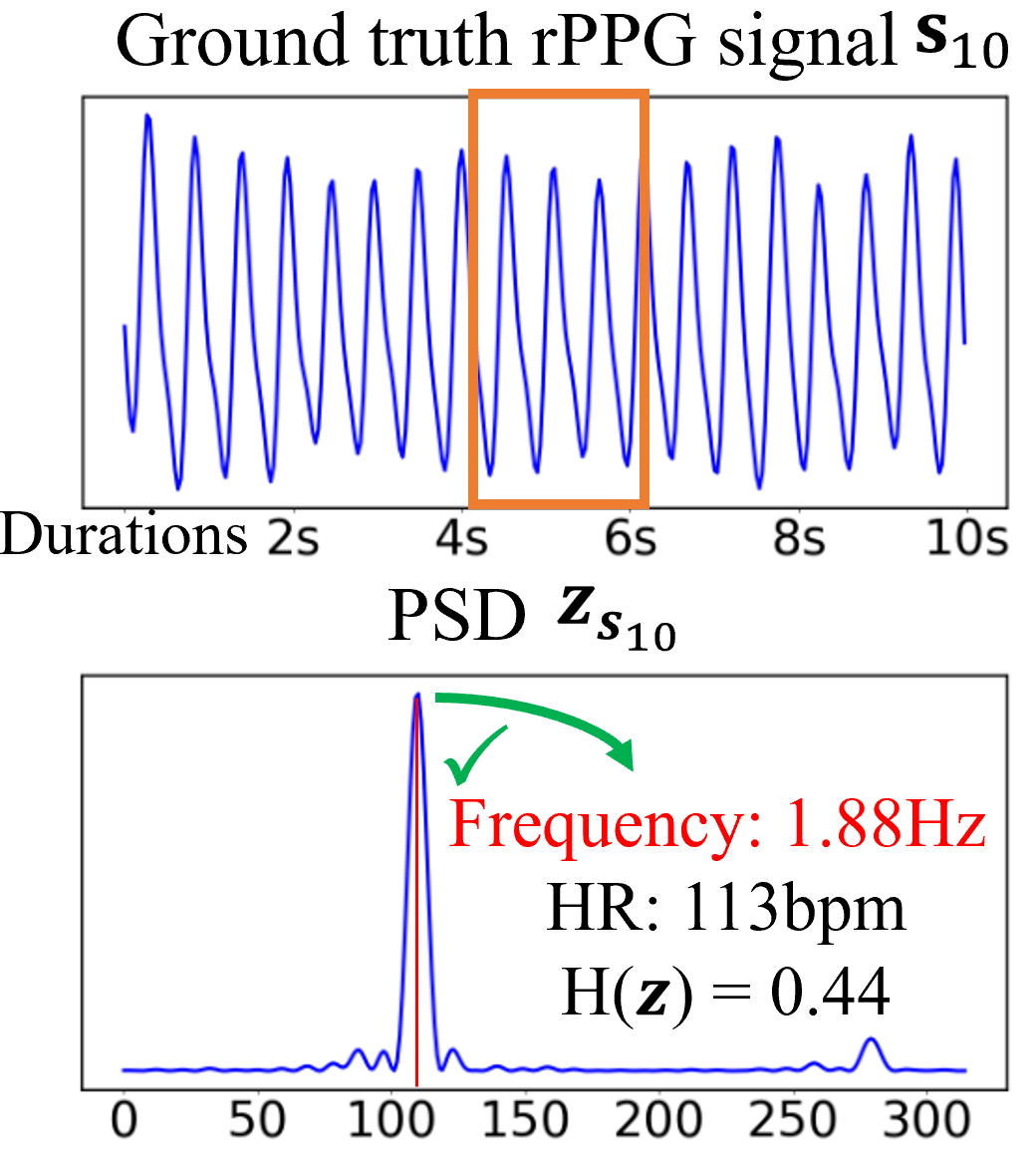}} &
    \fbox{\includegraphics[width=0.38 \columnwidth]{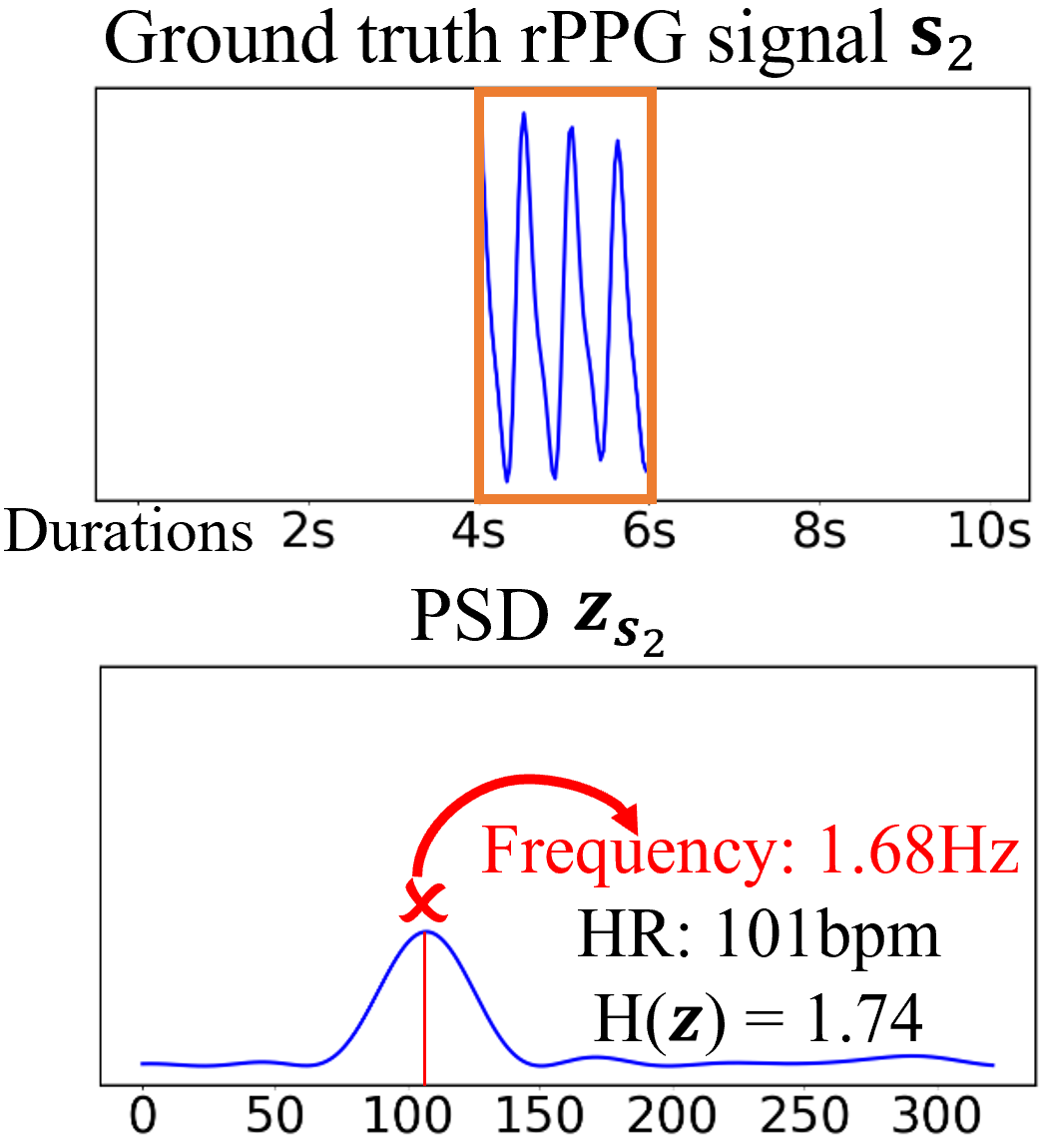}} 
    \\ (a)&(b) 
    \end{tabular} 
   
\caption{   
\textcolor{black}{Examples of rPPG signals and their corresponding PSDs:} (a) the ground truth 10-second rPPG signal, and (b) the ground truth 2-second rPPG signal.
}
\label{fig:issue_in_GT}   
\end{figure}

Comparing with measuring HRs from longer video clips, measuring HRs from ultra-short video clips faces significantly greater challenges.
The first challenge comes from the limited number of heartbeat cycles available in ultra-short video clips during inference stage. 
Since a full heartbeat cycle in rPPG signals lasts at least $ \frac{60}{40}=1.5$ seconds, \textcolor{black}{a 2-second clip} can capture only a few or just one complete heartbeat cycle. 
The second challenge stems from the spectral leakage issue  \cite{xiao2023harmonic, li2020frequency, liu2022learning} in ultra-short signals. As shown in Figure~\ref{fig:issue_in_GT}, spectral leakage often leads to imprecise PSDs in the estimated rPPG signals from ultra-short videos and consequently results in inaccurate HR measurements.

In this paper, we address the two key challenges in accurate HR measurement from ultra-short 2-second video clips. Our main idea is illustrated in Figure~\ref{fig:idea}.
First, since HR remains relatively stable over short periods \cite{sun2022contrast,gideon2021way}, we assume that rPPG signals estimated from different durations of the same video should exhibit consistent periodicity. Based on this insight, we propose an effective periodicity-guided rPPG estimation method that incorporates two loss terms: weighted spectral-based cross-entropy loss and maximized periodic similarity loss. These loss terms work together to mitigate spectral leakage and enforce consistent periodicity between rPPG signals estimated from ultra-short clips and their corresponding ground truth signals.
To further address the spectral leakage issue, we propose a novel periodicity-guided signal reconstruction by incorporating a generator to refine the rPPG model.
In particular, we train the generator to produce latent rPPG features corresponding to signals of varying lengths from ultra-short rPPG signals for reconstructing 10-second rPPG signals while maintaining their periodic consistency. 
The reconstructed longer rPPG signals are then used to enhance the rPPG model training for deriving more precise PSD estimation and accurate HR measurement.
Finally, we adopt an alternative optimization strategy to jointly train the rPPG model and the generator.
Our extensive experiments on intra-domain and cross-domain testing across four public databases demonstrate that the proposed method effectively measures HR with high accuracy from ultra-short video clips.

Our contributions are summarized as follows.

\begin{itemize}
\item 

We propose a novel periodicity-guided method to address two key challenges in heart rate (HR) measurement from ultra-short video clips: insufficient heartbeat cycles in ultra-short clips and HR estimation inaccuracy due to spectral leakage.

\item  
To address the issue of insufficient heartbeat cycles, we propose an effective periodicity-guided rPPG estimation method that enforces consistent periodicity between rPPG signals estimated from different durations of the same video.

\item 
To address the spectral leakage issue, we incorporate a generator to reconstruct longer rPPG signals from ultra-short ones while preserving their periodic consistency to enable more precise PSD estimation and accurate HR measurement in the rPPG model.

\item 
Our extensive experimental results demonstrate that the proposed method effectively measure HR from ultra-short clips and outperforms previous rPPG estimation methods to achieve state-of-the-art performance.  
 
\end{itemize}

\section{Related Work}  

\subsection{Remote heart rate measurement}
Remote heart rate (HR) measurement focuses on estimating remote photoplethysmography (rPPG) signals from video clips to measure HRs.  

Most recent supervised methods \cite{hsieh2022augmentation,yu2022physformer,du2023dual} proposed using paired ground truth rPPG signals and videos to learn rPPG estimation. 
In particular, in \cite{hsieh2022augmentation}, we proposed to augment rPPG datasets to enhance rPPG estimation by removing existing rPPG signals from videos and embeding new rPPG signals into videos.
Next, in \cite{yu2022physformer}, the authors proposed including Temporal Difference Convolution \cite{yu2020autohr} into transformer to  explore long-range spatiotemporal relationships for rPPG estimation.  
Furthermore, the authors in \cite{du2023dual} proposed synthesizing target noises to reduce the domain variations in rPPG estimation. 
In \cite{huang2026fully}, we focused on exploring the test-time adaptation (TTA) scenario for rPPG estimation to enhance the adaptation capability of pre-trained rPPG models.
In addition, some unsupervised methods  \cite{gideon2021way,sun2022contrast,huang2025dd} proposed using different characteristics of rPPG signals, including rPPG spatial similarity, rPPG temporal similarity,  cross-video rPPG dissimilarity, and HR range constraint to learn rPPG estimation. 
For example, in our previous work \cite{huang2025dd}, we proposed modeling interference features to help rPPG models derive de-interfered representations for learning genuine rPPG signals.
However, previous methods focused on learning rPPG estimation from long video clips, (\eg, 10-second video clips) to measure HRs. 
Remote HR measurement from ultra-short videos still remains unexplored in previous methods.
 
Although some existing methods \cite{huang2021novel,ouzar2023x} have attempted to predict HR from short video clips (e.g., 2-second clips). The authors in \cite{huang2021novel} combined 3D convolutional networks and LSTM to estimate pulse rate.
The authors in \cite{ouzar2023x} utilized depthwise separable 3D convolutions to enhance spatiotemporal feature extraction across color channels. 
Both methods employ MLP to directly regress HR values. However, as noted in \cite{liu2023information}, this process lacks mathematical interpretability regarding the relationship between the rPPG signal and HR values.
   
\subsection{Spectral leakage}

Spectral leakage in signal processing and Fourier analysis occurs when a signal is not perfectly periodic within the observation window, causing energy to "leak" into adjacent frequencies in the frequency spectrum and resulting in imprecise power spectral density (PSD).
To address this issue, previous methods generally fall into two  categories: window-based \cite{carbone2001windows, jwo2021windowing} and data-based \cite{xiao2023harmonic,li2020frequency} methods.
In \cite{carbone2001windows, jwo2021windowing}, the authors proposed to select  suitable windows to obtain precise power spectral density. 
In  contrary to window-based methods \cite{carbone2001windows, jwo2021windowing}, the authors in \cite{li2020frequency,xiao2023harmonic} proposed extending the data length while maintaining the same frequency to mitigate spectral leakage. 
In particular, in \cite{xiao2023harmonic}, the authors adopted data extrapolation on the basis of the sampling sequence to improve the frequency resolution. 
The authors in \cite{li2020frequency} proposed to use autoregressive models to increase the data length while maintaining the same sampling frequency  to mitigate spectral leakage. 

\begin{figure*}[t] 
\centering
    \begin{tabular}{@{}c@{}}
            \includegraphics[width=0.9 \linewidth]{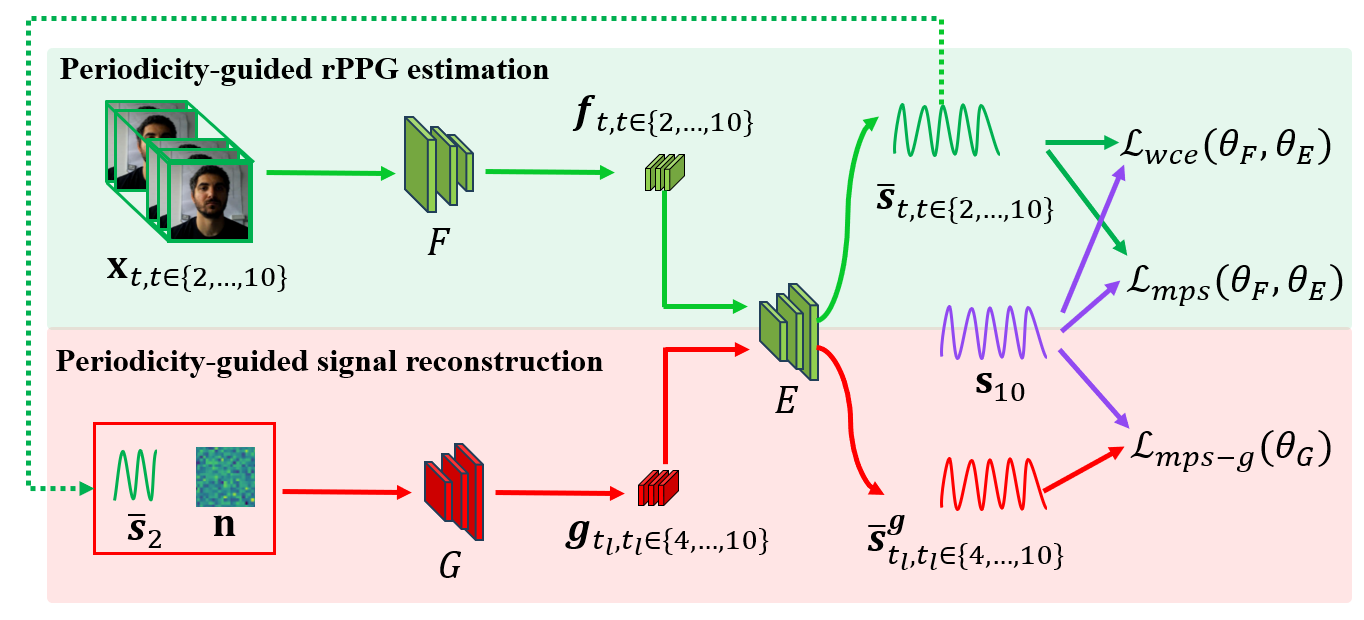} 
    \end{tabular}
    \caption{ The proposed periodicity-guided rPPG estimation and signal reconstruction for ultra-short rPPG estimation.  
    } 
    \label{fig:framework}   
\end{figure*}

\section{Proposed Method}
\label{sec:Proposed_method}

\subsection{Overview}
\label{sec:Overview}
As mentioned in the Introduction, the two key challenges (insufficient heartbeat cycles and spectral leakage) severely hinder HR estimation accuracy. To further illustrate these issues, we present an example in Figure~\ref{fig:issue_in_GT}.
As shown in Figures~\ref{fig:issue_in_GT} (a) and (b), the estimated power spectral densities (PSDs) from ground truth rPPG signals of 10-second and 2-second durations exhibit a noticeable frequency discrepancy. The frequency shift observed in Figure~\ref{fig:issue_in_GT} (b), caused by spectral leakage, deviates from the ground truth in the longer signal shown in Figure~\ref{fig:issue_in_GT} (a). This inaccuracy in PSD estimation ultimately leads to errors in HR measurement.

To overcome these challenges, we propose a novel \textbf{periodicity-guided rPPG estimation and signal reconstruction method}, as shown 
in Figure~\ref{fig:framework}. 
\textcolor{black}{
In \textbf{periodicity-guided rPPG estimation}, we first introduce a weighted spectral-based cross-entropy loss $\mathcal{L}_{wce}$ to alleviate the spectral leakage issue in ultra-short rPPG signals by assigning lower weights to unreliable frequency components with higher spectral entropy. Next, by leveraging the consistent periodic characteristics of rPPG signals, we propose a maximized periodic similarity loss $\mathcal{L}_{mps}$ to encourage the estimated rPPG signals to maintain the same underlying periodicity as their corresponding 10-second ground-truth rPPG signals.
In \textbf{periodicity-guided signal reconstruction},  we further leverage the consistent periodic characteristics of rPPG signals and introduce a generative loss $\mathcal{L}_{mps\text{-}g}$ to reconstruct the estimated 2-second rPPG signals into a continuous 10-second rPPG signal, while preserving the same underlying periodicity as the corresponding estimated 2-second rPPG signals.}
In the following sections, we provide detailed description of our approach.

\subsection{Periodicity-guided rPPG estimation} 
\label{sec:Periodicity-guided rPPG estimation}

Given the labeled training set $\{\mathbf{X},\mathbf{S}\}$, where $\mathbf{x} \in \mathbf{X}$ represents facial videos and $\mathbf{s} \in \mathbf{S}$ denotes their corresponding ground truth rPPG signals, we first divide each $\mathbf{x}$ and $\mathbf{s}$ into sets of clips $\mathbf{C}_t = \{\mathbf{x}_{t}\}$ and their corresponding sets of rPPG signals $ \mathbf{S}_t = \{\mathbf{s}_{t}\}$ of varying durations $t\in \{2,4,\cdots,10\}$, respectively.

Next, we introduce a spectral-based cross entropy loss $\mathcal{L}_{ce}$ to train the feature extractor $F$ and the rPPG estimator $E$ in $T=E\circ F$.
As shown in Figure~\ref{fig:framework}, let $\bar{\mathbf{s}}_t$ denote the rPPG signal estimated from $ {\mathbf{x}}_t$  by,
\begin{equation}
    \bar{\mathbf{s}}_t = T(\mathbf{x}_t) = E(F(\mathbf{x}_t)) = E(\mathbf{f}_t)\,, 
\label{eqn:modelT}
\end{equation}

\noindent 
where $\mathbf{f}_t = F(\mathbf{x}_t)$ denotes the latent rPPG features. 
{Following existing methods \cite{gideon2021way,sun2022contrast}, we identify the highest peak frequencies within $\mathbf{z}_{\bar{\mathbf{s}}_t}$ and $\mathbf{z}_{\mathbf{s}_t}$ to define the spectral-based cross-entropy loss $\mathcal{L}_{ce}$ as follows:}
\begin{equation}
\mathcal{L}_{ce}({\theta}_F, {\theta}_E) 
=  -\sum\nolimits_{i=a}^b  {\mathbf{z}_{\bar{\mathbf{s}}_t}^i}\log(\mathbf{z}_{\mathbf{s}_t}^i),
\label{eqn:CE}
\end{equation} 

\noindent
{where $\mathbf{z}_{\bar{\mathbf{s}}_t}$ and $\mathbf{z}_{ {\mathbf{s}}_t}$  are the estimated rPPG signal $\bar{\mathbf{s}}_t$ and the ground truth rPPG signal $\mathbf{s}_t$, respectively, and  $a$ and $b$ denote the lower and upper frequency bounds (\ie, 0.66 Hz and 4.16 Hz) for human HR \cite{speth2023non}.   }

\subsubsection{Weighted spectral-based cross entropy loss}

{In our previous study \cite{huang2026fully}, we show that the entropy of the PSDs of the predicted rPPG signals can effectively reflect their estimation accuracy. This property can therefore be leveraged to guide the rPPG model toward more reliable signal estimation.
}

However, ultra-short signals tend to yield inaccurate spectral estimation when computing PSDs due to spectral leakage.
As shown in Figure~\ref{fig:issue_in_GT} (b), the PSD derived from short rPPG signals differ significantly from that of the original longer signals in Figure~\ref{fig:issue_in_GT} (a) and exhibits higher entropy H({z}). 
Therefore, the rPPG model may learn inaccurate rPPG estimation from directly applying the spectral-based cross-entropy  constraint in Eq. \eqref{eqn:CE}. \textcolor{black}{To mitigate this issue,}
we propose a weighted spectral-based cross entropy loss $\mathcal{L}_{wce}$ to replace the  loss $\mathcal{L}_{ce}$ in \eqref{eqn:CE} by assigning smaller weights to frequency components with higher entropy as follows:
\begin{equation}
\mathcal{L}_{wce}({\theta}_F, {\theta}_E) 
= w_{\mathbf{s}_t} \cdot \mathcal{L}_{ce}({\theta}_F, {\theta}_E),
\label{eqn:W-CE}
\end{equation}

\noindent 
where the weight $ w_{\mathbf{s}_t}$ is determined in terms of the entropy  $\text{H}(\mathbf{z} _{\mathbf{s}_t})=-\sum\nolimits_{i=a}^b  {\mathbf{z}_{\mathbf{s}_t}^i}\log(\mathbf{z}_{\mathbf{s}_t}^i)$ of $\mathbf{s}_t$ by,
\begin{equation} 
w_{\mathbf{s}_t} =1-\frac{\text{H}({\mathbf{z} _{\mathbf{s}_t}})-\text{min}(W)}{\text{max}(W)-\text{min}(W)},
\label{eq:ent_w}
\end{equation}
  
\noindent  
and $W =\{\text{H}(\mathbf{z} _{\mathbf{s}_t})\}$ represents the set of weights derived from the entropy estimates of the PSDs for each ground truth rPPG signal $\mathbf{s}_{t}$.  
 
\subsubsection{Periodic characteristic of rPPG signals}
 
Nevertheless, while the weighted spectral-based cross entropy loss $\mathcal{L}_{wce}$ helps mitigate the impact of imprecise PSDs, this loss alone does not guarantee that the estimated rPPG signals $\bar{\mathbf{s}}_t$ maintain the same periodicity as the 10-second ground truth rPPG signals $\mathbf{s}_{10}$. 
To address this issue, we build on the observation from \cite{sun2022contrast} that rPPG signals of different durations within the same video should exhibit the same periodicity and frequency. 
\textcolor{black}{
To validate this assumption, in Table~\ref{tab:variations-HR}, we further use the ground truth rPPG signals to analyze the heart rate variations across clips of different durations from the same video in multiple public rPPG datasets.
As shown in Table~\ref{tab:variations-HR}, the observed heart-rate variations among clips extracted from the same video are generally very small. In addition, the ground truth rPPG signals from different clips within the same video exhibit consistently high Pearson Correlation Coefficient (R) values. These observations indicate that the ground truth rPPG signals of different durations  from the same video share highly similar periodicity and frequency characteristics.}
Based on this assumption, we further constrain the rPPG model by enforcing consistent rPPG periodicity across video clips of varying durations.

\begin{table}[t]
\centering
\small
\setlength\tabcolsep{2 pt} 
\caption{\textcolor{black}{Statistics of heart rate variations across temporal clips from the same video in public rPPG datasets.}}
\label{tab:variations-HR}
\color{black}
\scalebox{0.9}{
\begin{tabular}{ ccc|ccc|ccc|ccc }
\hline
   \multicolumn{3}{c|}{\textbf{U}}                                                                       & \multicolumn{3}{c|}{\textbf{P}}                                                                       & \multicolumn{3}{c|}{\textbf{C}}                                                                       & \multicolumn{3}{c }{\textbf{V}}                                                                       \\ \cline{1-12} 
 \multicolumn{1}{c|}{MAE$\downarrow$} & \multicolumn{1}{c|}{RMSE$\downarrow$} & R$\uparrow$  & \multicolumn{1}{c|}{MAE$\downarrow$} & \multicolumn{1}{c|}{RMSE$\downarrow$} & R$\uparrow$  & \multicolumn{1}{c|}{MAE$\downarrow$} & \multicolumn{1}{c|}{RMSE$\downarrow$} & R$\uparrow$  & \multicolumn{1}{c|}{MAE$\downarrow$} & \multicolumn{1}{c|}{RMSE$\downarrow$} & R$\uparrow$ \\ \hline
 
  \multicolumn{1}{c|}{0.09}           & \multicolumn{1}{c|}{0.50}            & 0.99          & \multicolumn{1}{c|}{0.09}           & \multicolumn{1}{c|}{0.38}            & 0.99          & \multicolumn{1}{c|}{0.19}           & \multicolumn{1}{c|}{1.47}            & 0.99          & \multicolumn{1}{c|}{0.15}           & \multicolumn{1}{c|}{0.93}            & 0.99      
 
 \\  
 \hline
\end{tabular} }
\end{table}

\paragraph{Differences between Classical NCC and the Proposed SWM-NCC Operations}

We first consider using the classical Normalized Cross-Correlation (NCC) operation to verify the consistency of rPPG periodicity across video clips with varying durations. However, when handling signals of unequal lengths, the classical NCC operation pads the shorter signal to match the longer one. Since the obtained similarity is derived from the zero-padded shorter signal rather than the original signal, this padding operation inevitably interferes with the similarity computation.
\textcolor{black}{
In contrast, instead of padding the shorter signal, we propose a Sliding Window Maximum Normalized Cross-Correlation (SWM-NCC) operation.
Specifically, the shorter signal $\mathbf{s}_{t}[v]$  is treated as a sliding window that progressively matches the longer signal $\hat{\mathbf{s}}_{t}[u]$ , and the maximum correlation is recorded to form the running correlation  $\mathbf{m}_{\mathbf{s}_{t},\hat{\mathbf{s}}_{t}}[\tau]$. This design enables a more effective characterization of rPPG periodic consistency across signals with varying durations:}
\begin{equation}
\label{eq:SWM-NCC}
\textcolor{black}{
\mathbf{m}_{\mathbf{s}_{t},\hat{\mathbf{s}}_{t}}[\tau] = \operatorname{SWM-NCC}(\mathbf{s}_{t}[v],\hat{\mathbf{s}}_{t}[u])  
= \max_{k} \text{NC} ( {\mathbf{s}}_{t}[v-k], \mathbf{s}_\tau[n])  
=  \max_{k}  \mathbf{c}_{\mathbf{s}_{t},\mathbf{s}_{\tau}}[k],
}
\end{equation}
\noindent 
{where $ L_{\mathbf{s}_t} \geq 1 $ and $ L_{\hat{\mathbf{s}}_t} \geq 1$ are the lengths of the
shorter signal $\mathbf{s}_{t}[v]$ and the
 longer signal $\hat{\mathbf{s}}_{t}[u]$, respectively,} $ L_{\mathbf{s}_t} \leq  L_{\hat{\mathbf{s}}_t}$,  $\text{NC}(a[v],b[v]) = \frac{\sum_l^L a[l] \cdot b[l]}{\sqrt{\sum_l^L (a[l])^2  }\sqrt{\sum_l^L (b[l])^2  }}$ is the normalized correlation between signals  $a[v]$ and $b[v]$,  $k \in \{-(L_{\mathbf{s}_t}-1),\cdots,0,\cdots,L_{\mathbf{s}_t}-1\}$ is the running lag for shifting the signal, 
$\mathbf{c}_{\mathbf{s}_{t},\mathbf{s}_{\tau}}$ denotes the running correlation calculated by the Normalized  Cross-Correlation (NCC) between $\mathbf{s}_{t}$ and  $\mathbf{s}_{\tau}$,
$\mathbf{s}_{\tau}$ denotes the smaller segments divided from the longer rPPG signal $\hat{\mathbf{s}}_{t}$, each having the same duration as the shorter rPPG signal $\mathbf{s}_{t}$, and is defined as follows:  
\begin{equation}
\begin{aligned}
    \mathbf{s}_\tau[n] =
    \begin{cases}
        \hat{\mathbf{s}}_t[n+\tau], & \text{if } 0 \leq n \leq L_{\mathbf{s}_t}-1, \\
        0, & \text{otherwise},
    \end{cases}
\end{aligned}
\end{equation}

\noindent 
where $\tau \in \{0, \ldots, L_{\hat{\mathbf{s}}_{t}} - L_{\mathbf{s}_t}\}$ indicates the running lag. 

\begin{figure}  
    \centering
     \centering 
    \begin{tabular}{c c} 
    \fbox{\includegraphics[width=0.385 \columnwidth]{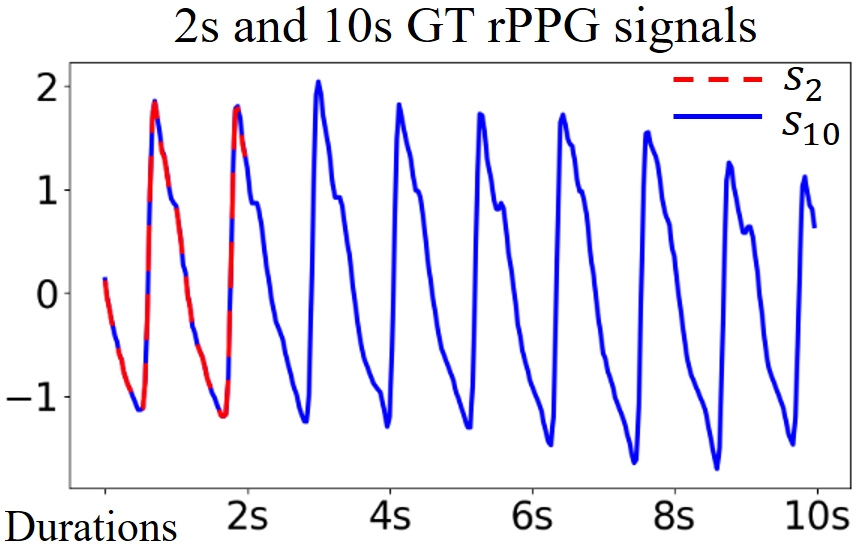}}  & 
       \fbox{\includegraphics[width=0.42\columnwidth ]{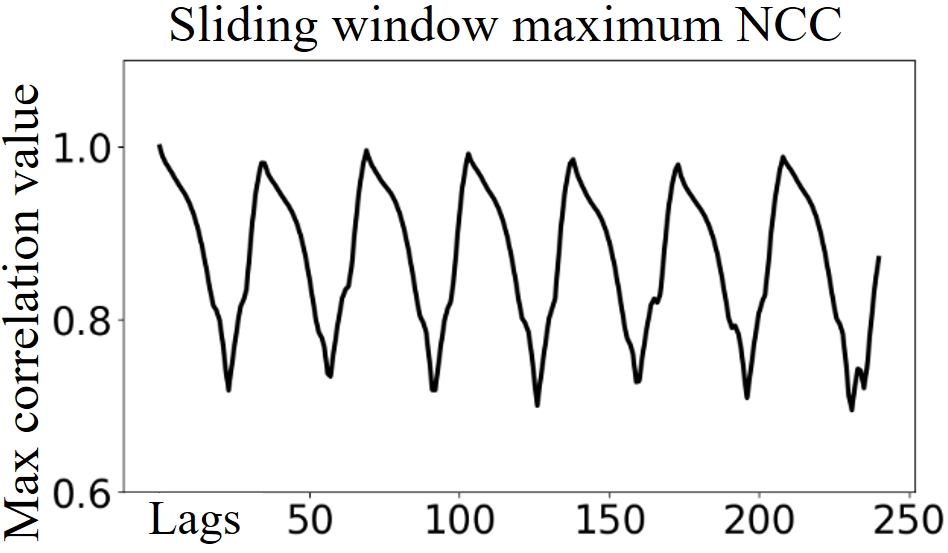}} 
    \\
     (a)  & (b) 
  \end{tabular}
\caption{  
\textcolor{black}{Examples of periodic consistency between short- and long-duration rPPG signals:} (a)	Examples of a 2-second rPPG signal ($\mathbf{s}_{2}$, in red) and its corresponding 10-second ground truth rPPG signal ($\mathbf{s}_{10}$, in blue l); and (b) their running correlation computed using SWM-NCC.  
}
\label{fig:observation} 
\end{figure}  
 
In Figure~\ref{fig:observation}, we present an example using the proposed SWM-NCC to show the running correlation $\mathbf{m}[\tau]$ between a 2-second rPPG signal $\mathbf{s}_{2}$ and its corresponding 10-second ground truth rPPG signal $\mathbf{s}_{10}$. As shown in Figure~\ref{fig:observation} (b), the high maximum running correlation indicates that rPPG signals of different durations from the same video indeed exhibit the same periodicity characteristics.

\subsubsection{Maximized periodic similarity loss}

Leveraging the consistent characteristics of rPPG periodicity, we define the maximized periodic similarity loss $\mathcal{L}_{mps}$ to ensure that the estimated rPPG signal $\bar{\mathbf{s}}_{t}$ maintains the same periodicity as the 10-second ground truth rPPG signal $\mathbf{s}_{10}$ from the same video. 
Since SWM-NCC($\cdot$) in \eqref{eq:SWM-NCC} is able to compute the running correlation between two signals to reflect their periodic consistency across varying durations, we now define the maximized periodic similarity loss $\mathcal{L}_{mps}$ to maintain the maximum correlation between the estimated rPPG signal $\bar{\mathbf{s}}_{t}$ and the 10-second ground truth rPPG signal $\mathbf{s}_{10}$  across {different heartbeat cycles} by,  
\begin{equation} 
\label{eqn:MPS loss}
\textcolor{black}{
\mathcal{L}_{mps} ({\theta}_F, {\theta}_E)   
={ 1 - \text{FP}( \mathbf{m}_{\bar{\mathbf{s}}_{t},{\mathbf{s}}_{10}}[\tau],\Delta_t )}
 = { 1 -  \frac{1}{N_{\mathbf{m}_h}}  \sum_{h=0}^{N_{\mathbf{m}_h}-1} \text{max}(\mathbf{m}_h[q]), } }
\end{equation}  

\noindent 
{where FP($\cdot$)  calculates the maximum correlation value between two signals across different  heartbeat cycles, 
$\mathbf{m}_{\bar{\mathbf{s}}_{t},{\mathbf{s}}_{10}}[\tau]=\text{SWM-NCC} (\bar{\mathbf{s}}_{t}[v], {\mathbf{s}}_{10}[u])$ denotes the running correlation between $\bar{\mathbf{s}}_{t} $ and $ {\mathbf{s}}_{10}$ obtained by SWM-NCC($\cdot$) in \eqref{eq:SWM-NCC}, $\Delta_t \geq  \frac{60}{40}=1.5$ denotes the time interval needed to capture at least one complete heartbeat cycle \cite{sun2022contrast}, 
$\mathbf{m}_h[q]$ denotes the running correlation segments between two signals, each containing at least one complete heartbeat cycle,  divided from $\mathbf{m}_{\bar{\mathbf{s}}_{t},{\mathbf{s}}_{10}}[\tau]$ with length $L_{\mathbf{m}_h} = fps \times \Delta_t$, $fps$  denotes the frame rates of different datasets,} 
$N_{\mathbf{m}_h} = \max (\lfloor \frac{| \mathbf{m}_{\bar{\mathbf{s}}_{t},{\mathbf{s}}_{10}}[\tau] |}{L_{\mathbf{m}_h}} \rfloor ,1) $ is the number of  $\mathbf{m}_h[q]$,  
and $\mathbf{m}_h[q]$ is obtained by, 
\begin{equation}
\begin{split} 
    \mathbf{m}_h[q] &= 
    \begin{cases}
        \mathbf{m}_{\bar{\mathbf{s}}_{t},{\mathbf{s}}_{10}}[q + h \cdot L_{m_{h}}], & \text{if } 0 \leq q \leq L_{\mathbf{m}_h}  -1, \\
        0, & \text{otherwise},
    \end{cases} 
\end{split}
\label{eqn:m_h}
\end{equation} 

\noindent  
where $h \in \{0, \ldots, N_{\mathbf{m}_h} - 1\}$. 

To summarize the training of the rPPG model $T$, we include the weighted spectral-based cross-entropy loss $\mathcal{L}_{wce}$ and the maximized periodic similarity loss $\mathcal{L}_{mps}$ to define the objective function for training $T=E\circ F$:
\begin{equation}
    \btheta^*_F, \btheta^*_E = \argmin_{\btheta_F, \btheta_E} \cL_{wce}(\btheta_F, \btheta_E)+ \cL_{mps}(\btheta_F, \btheta_E).
\label{eqn:theta_T}
\end{equation}

\subsection{Periodicity-guided signal reconstruction}
\label{sec:signal_reconstruction}

In this subsection, to mitigate the spectral leakage issue, we propose a novel periodicity-guided signal reconstruction approach to further enhance the rPPG model $T$ towards more accurate HR measurement. 
As noted in \cite{xiao2023harmonic, li2020frequency}, spectral leakage can be alleviated by extending short signals to longer ones while maintaining the same frequency.
To achieve this, we introduce a generator $G$ to reconstruct longer rPPG signals from ultra-short ones while preserving their periodic consistency.
In particular, we propose a progressive generation of latent rPPG features corresponding to rPPG signals of varying lengths. The details of rPPG feature generation and signal reconstruction are described below.

\begin{figure}[t]  
    \centering
    \includegraphics[width=7 cm]{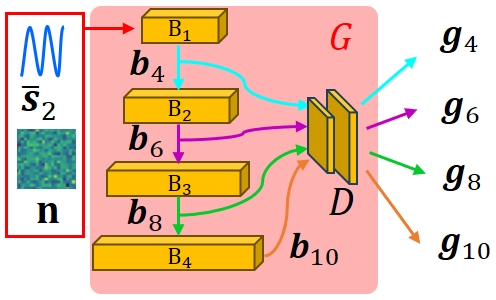} 
\caption{   
Progressive rPPG feature generation.
} 
    \label{fig:generator}
\end{figure} 

\subsubsection{Progressive feature generation}  

In the second part of Figure~\ref{fig:framework}, we illustrate how the generator $G$ is incorporated to reconstruct a set of rPPG signals $\bar{{\mathbf{s}}}_{t_l}^{g}$ , where $ t_l\in \{4, 6, 8,10\}$.
The details of the generator $G$ are given in Figure~\ref{fig:generator}, which consists of a set of hierarchical linear blocks $B_1$ - $B_4$ and a feature decoder $D$.
As shown in Figure~\ref{fig:generator}, given the estimated ultra-short rPPG signals $\bar{\mathbf{s}}_{2}$ from the rPPG model $T$, we use both $\bar{\mathbf{s}}_{2}$ and a unit Gaussian noise $\bn \sim \cN(0, I)$ as inputs to the generator $G$. 
The hierarchical linear blocks $B_1$ - $B_4$ in $G$ progressively extend the temporal dimensions to generate temporal block features $\bb_{t_l}$, which are decoded by $D$  to extend the spatial dimensions into latent rPPG features $\bg_{t_l}$ of varying durations $ t_l $ by, 
\begin{equation}
\begin{aligned}
{
     \mathbf{g}_{t_l} =  
    \begin{cases}
         D(B_{BN}(\bar{{\mathbf{s}}}_{2}), \mathbf{n}) =   D (\bb_{{t_l} }), & \text{if } BN= 1,  \\
           D(B_{BN}(\bb_{(t_l-2)} )) =  D (\bb_{{t_l}}), & \text{otherwise}.
    \end{cases}
    }
\end{aligned}
\label{eqn:generator}
\end{equation} 

\noindent 
where $BN= \frac{(t_l-2)}{2}$ denotes the index of the hierarchical linear blocks, and  $ t_l\in \{4, 6, 8,10\}$. 

\subsubsection{Periodicity constraint for signal reconstruction} 

To ensure periodic consistency between the reconstructed rPPG signals and the ground truth, we first obtain the corresponding rPPG signals $\bar{{\mathbf{s}}}_{t_l}^{g}  = E(\mathbf{g}_{t_l})$ from the generated rPPG features $\mathbf{g}_{t_l}$  for  $ t_l\in \{4, 6, 8,10\}$ using the estimator $E$. 

Next, we define the generative loss $\mathcal{L}_{mps\text{-}g}$ to constrain the generator $G$ for maintaining periodic consistency between  the reconstructed rPPG signals $\bar{{\mathbf{s}}}_{t_l}^{g}$  and their corresponding ground truth 10-second rPPG signal $ {\mathbf{s}}_{10}$ by,
\begin{equation}
\begin{aligned}
{
\mathcal{L}_{mps\text{-}g} (\theta_G) =  
\sum_{t_l} (1 - \text{FP}( \mathbf{m}_{\bar{\mathbf{s}}_{t_l}^g,{\mathbf{s}}_{10}}[\tau],\Delta_t ))  + 
\quad 
 \sum_{t_{\hat{l}}}
 (1 - \text{FP}( \mathbf{m}_{\bar{\mathbf{s}}^g_{t_{\hat{l}}},\bar{\mathbf{s}}^g_{t_{\hat{l}+2}} }[\tau],\Delta_t )), 
}
\end{aligned}
\label{eqn:G_NCC}
\end{equation} 

\noindent
{
where $ \mathbf{m}_{\bar{\mathbf{s}}_{t_l}^g,{\mathbf{s}}_{10}}[\tau]= \text{SWM-NCC} (\bar{\mathbf{s}}^g_{t_l}[v], \mathbf{s}_{10}[u])$ and $ \mathbf{m}_{\bar{\mathbf{s}}^g_{t_{\hat{l}}},\bar{\mathbf{s}}^g_{t_{\hat{l}+2}} }[\tau] = \text{SWM-NCC} (\bar{\mathbf{s}}^g_{t_{\hat{l}}}[v], \bar{\mathbf{s}}^g_{t_{\hat{l}+2}}[u])$ are the running
correlations between $\bar{\mathbf{s}}_{t_l}^g$ and $ {\mathbf{s}}_{10}$,  as well as between $\bar{\mathbf{s}}^g_{t_{\hat{l}}}$ and $\bar{\mathbf{s}}^g_{t_{\hat{l}+2}}$ obtained by SWM-NCC($\cdot$) in \eqref{eq:SWM-NCC},  $ t_{ {l}} \in \{ 4,6, 8, 10\}$, and $ t_{\hat{l}} \in \{ 4,6, 8\}$.}  
In \eqref{eqn:G_NCC}, the first term is designed to  learn the same periodicity characteristics between the reconstructed rPPG signals $\bar{{\mathbf{s}}}_{t_l}^{g}$ and the 10-second ground truth signals ${\mathbf{s}}_{10}$; 
and the second term  is designed to ensure the periodic consistency  among different reconstructed rPPG signals $\bar{{\mathbf{s}}}_{t_l}^{g}$.

Finally, the objective function for learning the parameters of $G$ is defined as follows:
\begin{equation}
    \btheta^*_G = \argmin\nolimits_{\btheta_G} \mathcal{L}_{mps\text{-}g} (\theta_G). 
\label{eqn:theta_G}
\end{equation} 

\subsection{Training and testing} 

\textcolor{black}{
We first adopt an alternative optimization strategy to train the rPPG model $T $ and the generator $G$ by iteratively solving the two coupled optimization problems of  \eqref{eqn:theta_G} and \eqref{eqn:theta_T}.  
In each iteration,  we first update  $F$ and  $E$ by minimizing $\cL_{wce}(\btheta_F, \btheta_E)$ in  \eqref{eqn:W-CE} and $ \cL_{mps}(\btheta_F, \btheta_E)$ in \eqref{eqn:MPS loss}. 
Next, we fix $F$ and $E$ and train $G$ by minimizing $\mathcal{L}_{mps\text{-}g} (\theta_G)$ in   \eqref{eqn:theta_G}.  
For inference, we first adopt $F$ and $E$ to  estimate the short rPPG signals $\bar{\mathbf{s}}_2$ from the ultra-short video clips ${\mathbf{x}}_2$.
Next, we adopt $G$ and $E$ to reconstruct long rPPG signals $\bar{\mathbf{s}}^{g}_{10}$ from $\bar{\mathbf{s}}_{2}$, and then  calculate the PSD of $\bar{\mathbf{s}}^{g}_{10}$ to derive HR.
}

  


\section{Experiments}

\subsection{Datasets} 

We conduct experiments on the following rPPG databases:  {UBFC-rPPG} \cite{bobbia2019unsupervised}  (denoted as \textbf{U}), {PURE} \cite{stricker2014non} (denoted as \textbf{P}), and {COHFACE}  \cite{heusch2017reproducible} (denoted as \textbf{C}), and { VIPL} \cite{niu2019rhythmnet} (denoted as \textbf{V}).
    
\paragraph{PURE}
\textcolor{black}{The PURE dataset \cite{stricker2014non} is one of the most widely used benchmark datasets for  rPPG estimation.
The heart-rate distribution of PURE is primarily concentrated in the ranges of 40–90 BPM and 115–140 BPM.
PURE comprises 60 RGB videos featuring 10 subjects. 
All videos are recorded in a resolution of 640 × 480 pixels at a frame rate of 30 fps. Each participant is recorded in six different scenarios, which include sitting still, speaking, slow and fast head movements, as well as small and large head rotations,  thereby introducing varying levels of motion interference. Each video has a duration of about one minute.
The videos are captured under relatively controlled indoor conditions with stable illumination and synchronized ground-truth pulse signals acquired from a contact-based pulse oximeter.  
}

\paragraph{COHFACE}
\textcolor{black}{The {COHFACE} dataset  \cite{heusch2017reproducible} dataset consists of 160 RGB videos of 40 subjects recorded under both studio and natural lighting conditions, introducing realistic illumination variations that are more challenging than those in PURE  \cite{stricker2014non}. 
The heart-rate distribution of COHFACE is primarily concentrated in the ranges of 40–90 BPM.
 In addition, subjects exhibit spontaneous facial expressions and natural head movements, resulting in diverse appearance and motion changes. 
 All videos are recorded in a resolution of 640 × 480 pixels and a frame rate of 20 fps,  which is notably lower than the 30 fps used in the other three datasets. 
Owing to its relatively unconstrained acquisition settings and synchronized physiological ground truth, COHFACE is commonly used to evaluate the robustness and generalization capability of rPPG methods in real-world scenarios.
}

\paragraph{UBFC-rPPG}
\textcolor{black}{
The UBFC-rPPG dataset \cite{bobbia2019unsupervised} is a widely used benchmark for remote heart-rate estimation under realistic recording conditions.
The heart-rate distribution of UBFC-rPPG is primarily concentrated in the ranges of 50–130 BPM.
The dataset {UBFC-rPPG}  contains  42 RGB videos captured under both indoor and natural lighting environments, introducing illumination variations commonly encountered in practical applications.
All videos are recorded in a resolution of 640 × 480 pixels with a frame rate of 30 fps.
During data collection, participants perform a time-sensitive mathematical task, which induces noticeable physiological and heart-rate fluctuations, making the dataset more challenging than those acquired under resting conditions. In addition, subjects exhibit natural facial movements and expression changes while solving the task, increasing the difficulty of rPPG estimation.
}

\paragraph{ VIPL}
\textcolor{black}{
The VIPL dataset \cite{niu2019rhythmnet} is one of the largest and most challenging public benchmarks for remote heart-rate estimation.
The heart-rate distribution of VIPL is primarily concentrated in the ranges of 50–180 BPM.
In particular, the {VIPL} dataset  \cite{niu2019rhythmnet}  includes 2,378 RGB videos of 107 subjects recorded under less-constrained conditions. Nine different conditions, including various head movements and illumination conditions are taken into consideration.
Unlike other rPPG datasets, VIPL-HR exhibits substantial variations in image quality and environmental conditions, making it more representative of real-world applications.
In addition, the VIPL dataset also includes recordings captured using both visible-light and near-infrared cameras, providing rich multimodal physiological data. 
Owing to its large scale, high diversity, and significant appearance variations, VIPL  is widely used to evaluate the robustness and generalization capability of rPPG methods.
Note that, although VIPL-HR provides both visible-light and near-infrared recordings, only the RGB modality is used in this work.
}


\subsection{Implementation details}

\subsubsection{Evaluation metrics} 
{To have a fair comparison with previous methods, we report all the results in terms of the following evaluation metrics: Mean Absolute Error (MAE) $\downarrow$, Root Mean Square Error (RMSE)  $\downarrow$, and Pearson Correlation Coefficient (R)   $\uparrow$.  }

\begin{figure*}[t] 
\centering
    \begin{tabular}{@{}c@{}}
            \includegraphics[width=1 \linewidth]{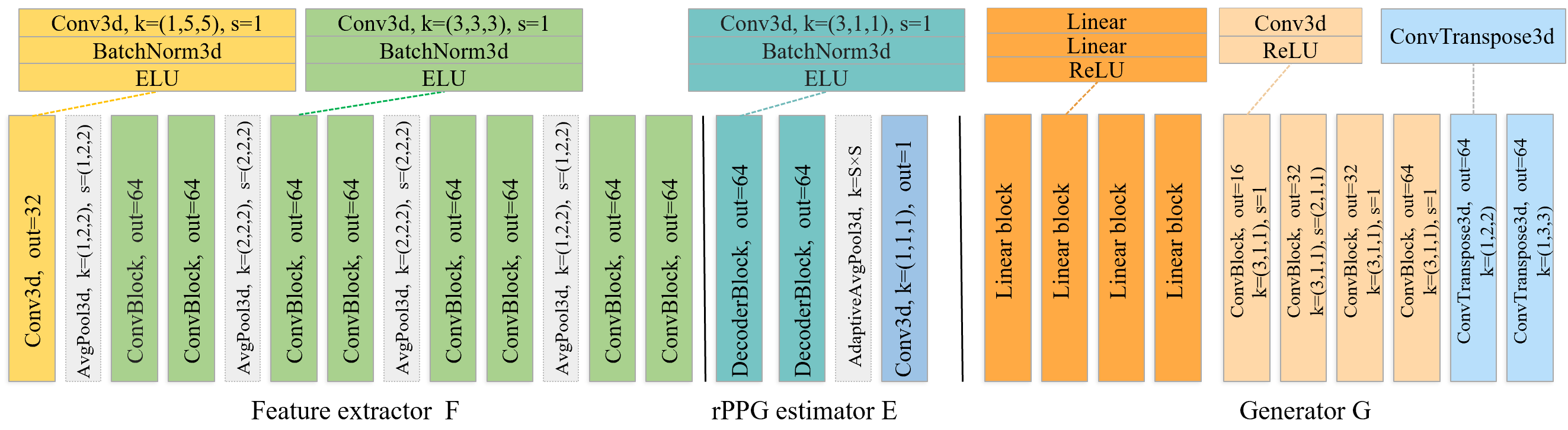} 
    \end{tabular}
    \caption{ 
    \textcolor{black}{Overall architecture of the proposed framework, consisting of the feature extractor $F$, the rPPG estimator $E$, and the signal generator $G$.}
    } 
    \label{fig:architecture}  
\end{figure*}

\subsubsection{Data preprocessing} 
\textcolor{black}{We use MTCNN~\cite{zhang2016joint} to detect facial regions, which are subsequently cropped and resized to a resolution of  64 $\times$ 64. In addition, to avoid initialization artifacts and obtain more stable physiological signals, we discard  the first 30 frames of each raw video. During the training stage, a 300-frame clip is randomly sampled from the remaining sequence of each preprocessed video for model training.
Each 10-second video is further divided into non-overlapping 2s, 4s, 6s, and 8s clips to construct training samples of different durations. For a fair comparison, both previous methods and the proposed method are retrained using either long video clips (e.g., 10s clips) or short video clips (e.g., 2s clips) as training data.
During the inference stage, each testing video is decomposed into non-overlapping 2-second clips for heart-rate estimation.
Due to the spectral leakage issue inherent in heart-rate estimation from ultra-short 2-second rPPG signals, the heart rate computed directly from the corresponding 2-second ground-truth rPPG segment may be unreliable, as illustrated in Figure~\ref{fig:issue_in_GT}. Therefore, for evaluation purposes, we derive the reference heart rate from the corresponding 10-second ground-truth rPPG signal that contains the target 2-second segment. The resulting heart-rate value is then used as the ground truth for MAE and RMSE computation.
Similarly, for the proposed method,  we compute the correlation coefficient (R) between the reconstructed 10-second rPPG signal and the corresponding 10-second ground-truth rPPG signal that contains the target 2-second segment.
}

\paragraph{Network architecture}
\textcolor{black}{
We use the network architecture  proposed in \cite{sun2022contrast}  for the rPPG model $T$.
Figure~\ref{fig:architecture} shows the detailed network architecture of the proposed framework, including the feature extractor $F$, the rPPG estimator $E$, and the signal generator $G$.
In particular,  $F$ consists of nine convolutional blocks. $E$ is composed of two decoder blocks followed by a convolutional layer for rPPG signal estimation. $G$ comprises four linear blocks, four convolutional blocks, and two convolutional layers for periodicity-guided signal reconstruction.
}

\subsubsection{Hyperparameter tuning}
\textcolor{black}{
We adopt the AdamW optimizer and use a constant learning rate of $1 \times 10^{-5}$ to train the rPPG model $T$ with a batch size of 6 for 100 epochs. Meanwhile, we use  the same optimization settings and hyperparameters to train 
 $G$.
During spectral computation, the heart-rate search range is restricted to 40--250 BPM to cover the typical physiological heart-rate range encountered in humans.
All experiments are implemented using PyTorch and performed on a single NVIDIA RTX 5090 GPU.
More implementation details are available at: \url{https://github.com/Pei-KaiHuang/ultra-short-rppg}
}

\begin{table}[h]
\centering
\footnotesize
\setlength\tabcolsep{1 pt} 
\caption{Comparison of intra-domain testing on \textbf{U},  \textbf{P}, \textbf{C}, and \textbf{V}, using 2-second testing clips.
}
\label{tab:intra_testing}
\scalebox{0.78}{
\begin{tabular}{ c|c|ccc|ccc|ccc|ccc }
\hline
\multirow{2}{*}{Types}                                                         & \multirow{2}{*}{Methods}         & \multicolumn{3}{c|}{\textbf{U}}                                                                       & \multicolumn{3}{c|}{\textbf{P}}                                                                       & \multicolumn{3}{c|}{\textbf{C}}                                                                       & \multicolumn{3}{c }{\textbf{V}}                                                                       \\ \cline{3-14} 
   &                                  & \multicolumn{1}{c|}{MAE$\downarrow$} & \multicolumn{1}{c|}{RMSE$\downarrow$} & R$\uparrow$  & \multicolumn{1}{c|}{MAE$\downarrow$} & \multicolumn{1}{c|}{RMSE$\downarrow$} & R$\uparrow$  & \multicolumn{1}{c|}{MAE$\downarrow$} & \multicolumn{1}{c|}{RMSE$\downarrow$} & R$\uparrow$  & \multicolumn{1}{c|}{MAE$\downarrow$} & \multicolumn{1}{c|}{RMSE$\downarrow$} & R$\uparrow$ \\ \hline
\multirow{2}{*}{\begin{tabular}[c]{@{}c@{}}Non- \\learning  \end{tabular}} & CHROM \cite{de2013robust} (TBE 2013)        & \multicolumn{1}{c|}{13.39}           & \multicolumn{1}{c|}{22.15}            & 0.31          & \multicolumn{1}{c|}{16.68}           & \multicolumn{1}{c|}{22.99}            & 0.32          & \multicolumn{1}{c|}{16.97}           & \multicolumn{1}{c|}{22.29}            & 0.18          & \multicolumn{1}{c|}{21.09}           & \multicolumn{1}{c|}{27.78}            & 0.16          \\  
 & POS \cite{wang2016algorithmic} (TBE 2016)          & \multicolumn{1}{c|}{14.06}           & \multicolumn{1}{c|}{23.26}            & 0.26          & \multicolumn{1}{c|}{15.73}           & \multicolumn{1}{c|}{22.33}            & 0.35          & \multicolumn{1}{c|}{18.18}           & \multicolumn{1}{c|}{24.01}            & 0.15          & \multicolumn{1}{c|}{21.71}           & \multicolumn{1}{c|}{28.88}            & 0.16          \\ \hline
\multirow{5}{*}{}                                                              & Contrast-Phys \cite{sun2022contrast} (ECCV 22)  & \multicolumn{1}{c|}{15.73}           & \multicolumn{1}{c|}{16.31}            & 0.36          & \multicolumn{1}{c|}{12.65}           & \multicolumn{1}{c|}{13.03}            & 0.41          & \multicolumn{1}{c|}{15.92}           & \multicolumn{1}{c|}{16.07}            & 0.29          & \multicolumn{1}{c|}{32.62}           & \multicolumn{1}{c|}{35.87}            & 0.08          \\ 
 & Gideon2021 \cite{gideon2021way}  (ICCV 21)     & \multicolumn{1}{c|}{14.47}           & \multicolumn{1}{c|}{15.30}            & 0.33          & \multicolumn{1}{c|}{12.88}           & \multicolumn{1}{c|}{17.04}            & 0.39          & \multicolumn{1}{c|}{17.79}           & \multicolumn{1}{c|}{18.93}            & 0.14          & \multicolumn{1}{c|}{33.11}           & \multicolumn{1}{c|}{35.57}            & 0.12          \\  
 & RErPPG-Net \cite{hsieh2022augmentation} (ECCV 22)    & \multicolumn{1}{c|}{9.00}            & \multicolumn{1}{c|}{10.09}            & 0.50          & \multicolumn{1}{c|}{6.25}            & \multicolumn{1}{c|}{7.83}             & 0.47          & \multicolumn{1}{c|}{11.06}           & \multicolumn{1}{c|}{11.85}            & 0.21          & \multicolumn{1}{c|}{29.65}           & \multicolumn{1}{c|}{33.09}            & 0.12          \\  
  & PhysFormer \cite{yu2022physformer}  (CVPR 22)    & \multicolumn{1}{c|}{7.38}            & \multicolumn{1}{c|}{9.63}             & 0.59          & \multicolumn{1}{c|}{6.56}            & \multicolumn{1}{c|}{8.29}             & 0.43          & \multicolumn{1}{c|}{11.87}           & \multicolumn{1}{c|}{13.63}            & 0.20          & \multicolumn{1}{c|}{25.89}           & \multicolumn{1}{c|}{27.90}            & 0.22          \\ 
  & Dual-bridging \cite{du2023dual} (CVPR 23) & \multicolumn{1}{c|}{7.51}            & \multicolumn{1}{c|}{8.67}             & 0.42          & \multicolumn{1}{c|}{5.82}            & \multicolumn{1}{c|}{6.74}             & 0.42          & \multicolumn{1}{c|}{14.98}           & \multicolumn{1}{c|}{16.17}            & 0.18          & \multicolumn{1}{c|}{27.35}           & \multicolumn{1}{c|}{29.85}            & 0.19          \\  
\multirow{2}{*}{\begin{tabular}[c]{@{}c@{}}Learning\\ based\end{tabular}}      & PRnet \cite{huang2021novel}  (BSPC 21)         & \multicolumn{1}{c|}{5.29}            & \multicolumn{1}{c|}{7.24}             & {\textbf{0.73}}          & \multicolumn{1}{c|}{4.94}            & \multicolumn{1}{c|}{5.44}             & 0.45          & \multicolumn{1}{c|}{9.24}            & \multicolumn{1}{c|}{9.44}             & 0.25          & \multicolumn{1}{c|}{16.81}           & \multicolumn{1}{c|}{17.08}            & 0.16          \\   
 & X-iPPGNet  \cite{ouzar2023x}  (CBM 23)      & \multicolumn{1}{c|}{4.99}            & \multicolumn{1}{c|}{6.26}             & 0.67          & \multicolumn{1}{c|}{4.61}            & \multicolumn{1}{c|}{5.36}             & 0.49          & \multicolumn{1}{c|}{8.89}            & \multicolumn{1}{c|}{9.16}             & 0.24          & \multicolumn{1}{c|}{14.71}           & \multicolumn{1}{c|}{15.36}            & 0.12          \\  
\multirow{5}{*}{}                                                              & EfficientPhys  \cite{liu2023efficientphys} (WACV 23) & \multicolumn{1}{c|}{11.70}           & \multicolumn{1}{c|}{20.00}            & 0.62          & \multicolumn{1}{c|}{9.03}            & \multicolumn{1}{c|}{16.95}            & 0.38          & \multicolumn{1}{c|}{28.83}           & \multicolumn{1}{c|}{37.12}            & 0.02          & \multicolumn{1}{c|}{29.00}               & \multicolumn{1}{c|}{36.74}                & 0.02             \\   
 & DD-rPPGNet\cite{huang2025dd} (TIFS 25)    & \multicolumn{1}{c|}{18.77}           & \multicolumn{1}{c|}{25.00}            & 0.28          & \multicolumn{1}{c|}{5.23}            & \multicolumn{1}{c|}{6.66}             & -0.02         & \multicolumn{1}{c|}{17.16}           & \multicolumn{1}{c|}{21.02}            & 0.03          & \multicolumn{1}{c|}{-}               & \multicolumn{1}{c|}{-}                & -             \\  
 \cline{2-14} 
 & Ours                             & \multicolumn{1}{c|}{{\textbf{4.29}}}            & \multicolumn{1}{c|}{{\textbf{5.25}}}             & 0.64          & \multicolumn{1}{c|}{\textbf{1.33}}            & \multicolumn{1}{c|}{{\textbf{1.97}}}             & {\textbf{0.67}}          & \multicolumn{1}{c|}{{\textbf{6.30}}}            & \multicolumn{1}{c|}{{\textbf{6.57}}}             & {\textbf{0.32}}          & \multicolumn{1}{c|}{{\textbf{9.39}}}            & \multicolumn{1}{c|}{{\textbf{10.42}}}            & {\textbf{0.33}}          \\ \hline
\end{tabular} }
\end{table}

\subsection{Intra-domain and cross-domain testing }
  
\textcolor{black}{In Tables~\ref{tab:intra_testing}, \ref{tab:cross_testing}, and \ref{tab:cross_testing_v},} we conduct intra-domain and cross-domain testing and compare the results with previous rPPG estimation methods. 
Note that, we also include short video clips to retrain previous methods for a fair comparison. 

\subsubsection{Intra-domain testing} 
Table~\ref{tab:intra_testing} shows the intra-domain testing results on \textbf{U}, \textbf{P}, \textbf{C}, and \textbf{V} by using  2-second testing video clips.
First, as for non-learning based methods (CHROM \cite{de2013robust} and POS \cite{wang2016algorithmic}), which rely on mathematical models to capture color changes related to blood flow for HR measurement, they may perform even worse on ultra-short video clips compared to some previous learning-based methods under intra-domain testing. 
Next, although some learning-based methods \cite{huang2021novel} \cite{ouzar2023x} perform rPPG estimation on short videos by using rPPG models to directly predict HR values instead of first estimating rPPG signals and then computing HR via PSD,  these approaches lack mathematical interpretability regarding the relationship between PPG signals and HR values, as noted in \cite{liu2023information}.
Furthermore, since unsupervised rPPG estimation methods such as Contrast-Phys \cite{sun2022contrast} and Gideon2021 \cite{gideon2021way} train rPPG models without using ground truth rPPG signals, their performance is poor compared to supervised rPPG estimation methods, especially under ultra-short rPPG estimation scenarios.
Moreover, because the datasets \textbf{C} and \textbf{V} exhibit significant illumination variations, they pose a greater challenge  for supervised rPPG estimation methods compared to datasets \textbf{U} and \textbf{P}.
Finally, even though the datasets \textbf{C} and \textbf{V} remains a significant challenge, the proposed method,  by maintaining the consistent periodicity  characteristic between the estimated ultra-short rPPG signals and the reconstructed rPPG signals, significantly improves performance and outperforms the other methods for HR measurement from ultra-short video clips.

\begin{table*}[h]
\centering
\small
\setlength\tabcolsep{1.5pt}
\renewcommand{\arraystretch}{1.0} 
\caption{Comparison of cross-domain testing on  \mytab{\textbf{P}$+$\textbf{C}$\rightarrow$\textbf{U}}, \mytab{\textbf{U}$+$\textbf{C}$\rightarrow$\textbf{P}}, and 
\mytab{\textbf{U}$+$\textbf{P}$\rightarrow$\textbf{C}}, using 2-second testing clips.} 
\label{tab:cross_testing}
\scalebox{0.75}{
\centering
\begin{tabular}{ccccccccccc}
\hline
\multirow{3}{*}{{Types}} 
    & \multirow{3}{*}{\mytab{Methods}} 
    & \mc{3}{\mytab{\textbf{P}$+$\textbf{C}$\rightarrow$\textbf{U}}}
    & \mc{3}{\mytab{\textbf{U}$+$\textbf{C}$\rightarrow$\textbf{P}}}
    & \mc{3}{\mytab{\textbf{U}$+$\textbf{P}$\rightarrow$\textbf{C}}}  \\
    \cmidrule{3-5} \cmidrule{6-8} \cmidrule{9-11}
    &
    & \mytab{MAE$\downarrow$}
    & \mytab{RMSE$\downarrow$}
    & \multirow{1}{*}{\mytab{R$\uparrow$}}
    & \mytab{MAE$\downarrow$}
    & \mytab{RMSE$\downarrow$}
    & \multirow{1}{*}{\mytab{R$\uparrow$}}
    & \mytab{MAE$\downarrow$}
    & \mytab{RMSE$\downarrow$}
    & \multirow{1}{*}{\mytab{R$\uparrow$}}\\
    \midrule
    \multirow{2}{*}{\mytab{Non-learning \\ based}}
     & CHROM \cite{de2013robust} (\textit{TBE 2013})  & 13.39 & 22.15 & 0.31 & 16.68 & 22.99 & 0.32 & 16.97 & 22.29 & 0.18  \\
    & POS \cite{wang2016algorithmic} (\textit{TBE 2016}) & 14.06 & 23.26 & 0.26 & 15.73 & 22.33 & 0.35 & 18.18 & 24.01 & 0.15   \\
    \addlinespace[1pt]
    \cline{1-11}
    \addlinespace[2pt]
    \multirow{7}{*}{\mytab{ \\ Learning \\ based}}
    & Contrast-Phys \cite{sun2022contrast} (\textit{ECCV 22}) & 23.27 & 23.58 & 0.31 & 15.86 & 17.75 & 0.21 & 21.50 & 23.02 & 0.29  \\
    & Gideon2021 \cite{gideon2021way} (\textit{ICCV 21}) & 26.28 & 27.16 & 0.18 & 15.42 & 17.11 & 0.10 & 22.76 & 23.36 & 0.18  \\
    & RErPPG-Net \cite{hsieh2022augmentation} (\textit{ECCV 22}) & 20.27 & 24.45 & 0.25 & 10.09 & 12.59 & 0.39 & 17.30 & 20.69 & 0.37  \\
    & PhysFormer \cite{yu2022physformer} (\textit{CVPR 22}) & 34.12 & 37.89 & 0.10 & 19.90 & 26.41 & 0.21 & 18.89 & 23.75 & 0.04  \\
    & Dual-bridging \cite{du2023dual} (\textit{CVPR 23}) & 26.92 & 28.09 & 0.30 & 9.41 & 11.89 & 0.20 & 16.22 & 17.56 & 0.18   \\
    & PRnet  \cite{huang2021novel} (\textit{BSPC 21}) & 15.09 & 15.30 & 0.32 & 9.60 & 10.02 & 0.19 & 17.96 & 18.12 & 0.10   \\

    & X-iPPGNet  \cite{ouzar2023x} (\textit{CBM 23}) & 14.40 & 14.93 & 0.34 & 9.53 & 9.76 & 0.31 & 16.01 & 16.25 & 0.27   \\
    & EfficientPhys \cite{liu2023efficientphys} \textit{(WACV 23)}
    & 24.23 & 32.51 & 0.15
    & 18.22 & 29.70 & 0.36
    & 24.48 & 34.37 & 0.08\\
    
    & DD-rPPGNet \cite{huang2025dd} \textit{(TIFS 25)} 
    & 22.88 & 28.42 & 0.00 
    & 14.06 & 21.57 & 0.23  
    & 19.41 & 23.64 & 0.00 \\
    
    
    \addlinespace[1pt]
    \cline{2-11}
    \addlinespace[2pt]
    
    & {Ours} &{\textbf{10.96}}  & {\textbf{11.38}} & {\textbf{0.37}} & {\textbf{8.43}} & {\textbf{8.57}} & {\textbf{0.44}} & {\textbf{13.46}} & {\textbf{16.02}} & {\textbf{0.41}}  \\
    \hline
\end{tabular}} 
\end{table*} 

\begin{table*}[t]
\centering
\small
\setlength\tabcolsep{1.5pt}
\renewcommand{\arraystretch}{1.0} 
\caption{
\textcolor{black}{Comparison of cross-domain testing on  \mytab{\textbf{V}$\rightarrow$\textbf{U}}, and 
\mytab{\textbf{V}$\rightarrow$\textbf{P}}, using 2-second testing clips.}
} 
\label{tab:cross_testing_v}
\color{black}
\scalebox{0.75}{
\centering
\begin{tabular}{ccccccc}
\hline
\multirow{3}{*}{\mytab{Methods}} 
    & \mc{3}{\mytab{\textbf{V}$\rightarrow$\textbf{U}}}
    & \mc{3}{\mytab{\textbf{V}$\rightarrow$\textbf{P}}} \\
    \cmidrule{2-4} \cmidrule{5-7}
    &
    \mytab{MAE$\downarrow$}
    & \mytab{RMSE$\downarrow$}
    & \multirow{1}{*}{\mytab{R$\uparrow$}}
    & \mytab{MAE$\downarrow$}
    & \mytab{RMSE$\downarrow$}
    & \multirow{1}{*}{\mytab{R$\uparrow$}}\\
    \midrule
   
    Contrast-Phys \cite{sun2022contrast} (\textit{ECCV 22}) 
    & 28.00 & 30.36 & 0.11 
    & 32.11 & 36.43 & 0.14   \\

    RErPPG-Net \cite{hsieh2022augmentation} (\textit{ECCV 22}) 
    & 17.60 & 18.37 & 0.23    
    & 10.87 & 13.13 & 0.22  \\
    PhysFormer \cite{yu2022physformer} (\textit{CVPR 22}) 
    & 28.16 & 32.01 & 0.10 
    & 16.27 & 21.74 & 0.14   \\
    PRnet  \cite{huang2021novel} (\textit{BSPC 21}) 
    & 25.85 & 25.96 & -0.10 
    & 33.54 & 33.55 & 0.13    \\
    EfficientPhys \cite{liu2023efficientphys} \textit{(WACV 23)}
    & 32.66 & 38.93 & 0.07
    & 28.68 & 37.99 & 0.05 \\

    RhythmFormer  \cite{zou2025rhythmformer} \textit{(PR 25)}
    & 24.66 & 28.66 & 0.09
    & 18.27 & 23.68 & 0.10 \\
    
    \addlinespace[1pt]
    \cline{1-7}
    \addlinespace[2pt]
    
    {Ours}
    & \textbf{15.14} & \textbf{16.17} & \textbf{0.37}
    & \textbf{9.99} & \textbf{12.27} & \textbf{0.31} \\
    \hline
\end{tabular}} 
\end{table*} 

\subsubsection{Cross-domain testing}  
In Table~\ref{tab:cross_testing},  we refer to \cite{chung2022domain} to conduct cross-domain testing on \textbf{U}, \textbf{P}, and \textbf{C} by using  2-second testing video clips.  
First,  since \textcolor{black}{non-learning-based methods} do not rely on predefined training data, they are not affected by the cross-domain issue, as noted in \cite{chung2022domain}.
Next, due to the cross-domain shift between the training and testing data, we observe that previous learning-based rPPG estimation methods exhibit significantly reduced performance when encountering an unseen domain.
In comparison, the proposed method, which learns the intrinsic periodicity characteristics of rPPG signals,  demonstrates improved domain generalization ability in effectively  measuring HR from ultra-short video clips even in cross-domain scenarios. 
From these cross-testing experiments, we confirm that the proposed method effectively estimates ultra-short rPPG signals and reconstructs long rPPG signals  while maintaining the same frequency, rather than merely fitting the training dataset distribution.
 
\textcolor{black}{In Table~\ref{tab:cross_testing_v}, we further use \textbf{V} as the source domain and \textbf{U} and \textbf{P} as the target domains to conduct cross-domain evaluation using 2-second testing video clips.
As shown in Table~\ref{tab:cross_testing_v}, we observe that other methods exhibit noticeable performance degradation due to the challenging domain gap between VIPL-HR and the target datasets. These domain gaps mainly arise from variations in illumination conditions, head poses, recording devices, and image quality. In contrast, the proposed method exhibits only a relatively small performance degradation and consistently achieves the best performance on both \textbf{V}$\rightarrow$\textbf{U} and \textbf{V}$\rightarrow$\textbf{P} protocols.
By learning the periodic characteristics of rPPG signals, the proposed method encourages the rPPG model to focus on intrinsic physiological patterns rather than dataset-specific appearance cues, thereby maintaining relatively stable performance when domain information varies significantly.
}

\subsection{Ablation study}

\subsubsection{On different loss terms} 

In Table~\ref{tab:ablation_loss_E}, to evaluate the effectiveness of the proposed periodicity-guided rPPG estimation in Section~\ref{sec:Periodicity-guided rPPG estimation}, we compare using different combinations of loss terms to train the rPPG models for estimating rPPG signals from ultra-short video clips, without incorporating the generator for long rPPG signal reconstruction, on the intra-testing datasets \textbf{P} and \textbf{U}.

\begin{table}[t]
\centering
\setlength{\tabcolsep}{0.2pt}
\caption{ 
Ablation study for rPPG model $T$ on the intra-testing datasets \textbf{P} and \textbf{U}, \textcolor{black}{ using 2-second testing clips and different loss combinations. }
} 
\label{tab:ablation_loss_E}
\setlength{\tabcolsep}{1.3pt} 
\small
\color{black}
\centering
\begin{tabular}{ >{\centering\arraybackslash}c c c c  c
                | >{\centering\arraybackslash}c c c
                | >{\centering\arraybackslash}c c c
                }
    \hline
    \multicolumn{5}{c|}{\multirow{1}{*}{{Loss Terms}}} & 
    \multicolumn{3}{c|}{\textbf{P}} & 
    \multicolumn{3}{c }{\textbf{U}}  
    \\
    \cline{1-11}
    $\mathcal{L}_{ce}$  & $\mathcal{L}_{wce}$ & $\mathcal{L}_{ncc}$ & $\mathcal{L}_{acf}$ & 
    $\mathcal{L}_{mps}$ &
    
    MAE$\downarrow$ & RMSE$\downarrow$ & 
    R$\uparrow$ & 
    MAE$\downarrow$ & RMSE$\downarrow$ & 
    R$\uparrow$ 
    \\\hline
     \checkmark & - & - & - &- & 5.51 & 5.96 & 0.11 &  7.21 & 8.57 & 0.20 \\
  
      - & \checkmark & - & -&- & 4.76 & 5.10 & 0.24 &  6.91 & 8.22 & 0.38 \\

    - & - & \checkmark & -&- & 4.53 & 4.99 & 0.37 &  6.41 & 7.38 & 0.30 \\
      
      - & - & - &- & \checkmark& 2.93 & 3.16 & 0.42 &  5.72 & 6.61 & 0.41 \\
    \checkmark  & - & \checkmark &- & -& 3.91 &  4.70 & 0.31 &  
    5.71 & 7.17 & 0.38 \\ 
     \checkmark  & - & - & \checkmark & -& 4.12 &  5.39 & 0.39 &  
    7.98 & 10.56 & 0.48 \\ 
    -& \checkmark & - &- &   \checkmark & \textbf{2.69} &  \textbf{3.14} & \textbf{0.43} &  
    \textbf{5.17} & \textbf{6.38} & \textbf{0.50}  \\ 
    \hline
\end{tabular}
\end{table}

First, by comparing the cases of $\mathcal{L}_{ce}$ in \eqref{eqn:CE} vs. $\mathcal{L}_{wce}$ in \eqref{eqn:W-CE}, we see that using a small weight for high entropy of PSDs in $\mathcal{L}_{wce}$ indeed reduces the impact of imprecise PSDs and improves the performance compared to the case with $\mathcal{L}_{ce}$.

{Next, we compare the case where $\mathcal{L}{mps}$ is based on the proposed SWM-NCC operation in \eqref{eqn:MPS loss} with the case based on the classical NCC operation, i.e., the off-the-shelf normalized correlation loss $\mathcal{L}_{ncc}(\theta_F, \theta_E) = 1 - \max_{\tau} \mathbf{c}_{\bar{s}_t, s_t} [\tau]$, where $\mathbf{c}_{\bar{s}_t, s_t} [\tau] = NC( \bar{s}_t [v], s_t [v - \tau] )$ denotes the measured running correlation.}
However, the off-the-shelf $\mathcal{L}_{ncc}$ only enforces the maximum correlation between two rPPG signals as a whole, rather than constraining the maximum correlation for each complete heartbeat cycle. 
We observe that the rPPG model $T$ fails to effectively estimate accurate ultra-short rPPG signals from ultra-short video clips, resulting in poor performance.
In contrast,  the proposed $\mathcal{L}_{mps}$ ensures that the maximum correlation between ultra-short rPPG signals and long ground truth rPPG signals is maintained for each complete heartbeat cycle, thereby effectively guiding the rPPG model to estimate accurate ultra-short rPPG signals.

\textcolor{black}{
Furthermore, we investigate the effectiveness of incorporating an Auto-Correlation Function (ACF)-based constraint $\mathcal{L}_{acf}(\theta_F, \theta_E)$  for periodicity modeling, where the Negative Pearson Loss is employed to enforce consistency between the autocorrelation values of the reconstructed rPPG signals and those of the corresponding ground-truth rPPG signals.
Specifically, we compare the performance of $\mathcal{L}_{ce}$+$\mathcal{L}_{ncc}$,  $\mathcal{L}_{ce}$+$\mathcal{L}_{acf}$, and  $\mathcal{L}_{wce}$+$\mathcal{L}_{mps}$.
The results show that our method outperforms both NCC-based and ACF-based alternatives. 
Unlike ACF, which measures periodicity within a single signal through autocorrelation analysis, the proposed SWM-NCC explicitly evaluates periodic consistency between ultra-short estimated rPPG signals and their reconstructed long-duration counterparts. Therefore, SWM-NCC is more closely aligned with the objective of the proposed ultra-short rPPG estimation framework.
In addition, ACF relies on repeated periodic patterns, its reliability decreases when only a limited number of cardiac cycles are available in an ultra-short observation window.
The superior performance achieved by $\mathcal{L}_{wce}$+$\mathcal{L}_{mps}$ further demonstrates the effectiveness of the proposed periodicity-guided rPPG estimation strategy.
}

\subsubsection{On different reconstructions} 
\begin{figure}[t]  
    \centering
     \begin{tabular}{ccc} 
    {\includegraphics[width=0.2\columnwidth]{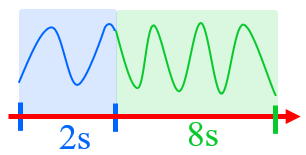}} &
    {\includegraphics[width=0.2\columnwidth]{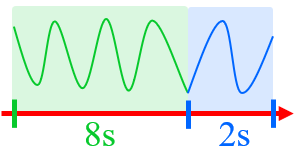}}&

    {\includegraphics[width=0.2\columnwidth]{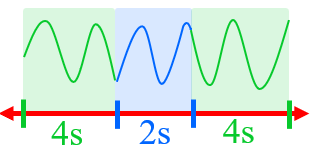}} 
    \\ (a)&(b)  &(c)  
    \end{tabular} 
  
\caption{   
Different rPPG reconstructions, including (a) forward, (b) backward, and (c) forward \& backward reconstructions (blue line: ground rPPG signals, green line: reconstructed rPPG signals).
}
\label{fig:reconstructions}  
\end{figure}

In Table~\ref{tab:G_frame}, we compare using different rPPG reconstructions to reconstruct the long rPPG signals.
Note that, w/o reconstruction indicates measuring HR using only the estimated ultra-short rPPG signals, and these results are consistent with the case of $\mathcal{L}_{wce}$ + $\mathcal{L}_{mps}$ in Table~\ref{tab:ablation_loss_E}.
Duplication refers to extending the estimated ultra-short rPPG signals by duplicating them to create long rPPG signals.
Forward, backward, and forward \& backward indicate different reconstructions for $G$, as shown in Figure~\ref{fig:reconstructions}.
 
Compared with w/o reconstruction, naive duplication fails to recover temporal information and degrades performance. We further compare different reconstruction strategies using the same loss $\mathcal{L}_{mps\text{-}g}$ and generator architecture (Figure~\ref{fig:reconstructions}). The forward, backward, and forward \& backward strategies all alleviate spectral leakage and improve PSD estimation over the no-reconstruction baseline. Among them, forward \& backward reconstruction achieves the most accurate rPPG signals by reducing the length of unidirectional missing segments. Therefore, forward \& backward reconstruction is adopted in all subsequent experiments.

\begin{table}[t]
\centering
\setlength{\tabcolsep}{1pt}
\caption{ 
Ablation study for generator $G$ on the intra-testing
datasets \textbf{P} and \textbf{U}, \textcolor{black}{ using 2-second testing clips and different reconstructions.}
} 
\label{tab:G_frame}
\small
\centering
\begin{tabular}{ >{\centering\arraybackslash}c 
                | >{\centering\arraybackslash}c c c
                | >{\centering\arraybackslash}c c c
                }
    \hline
    \multirow{1}{*}
    {{Different}}  & 
    \multicolumn{3}{c|}{\textbf{P}
    } & 
    \multicolumn{3}{c }{\textbf{U}} 
    \\
    \cline{2-7}
    {reconstructions}
      &
    MAE$\downarrow$ & RMSE$\downarrow$ & 
    R$\uparrow$ & 
    MAE$\downarrow$ & RMSE$\downarrow$ & 
    R$\uparrow$ 
    \\\hline
    w/o reconstruction  &  2.69 & 3.14 & 0.43 & {5.17} & {6.38} & {0.50}
       \\
        duplication&  2.90 & 3.23 & 0.25 & 4.80 & 5.76 & 0.39 \\
    forward  &  2.29 & 2.76 & 0.46 & 4.46 & 5.24 & 0.42
       \\
    backward  &  2.20 & 2.56 & 0.47 & 
    4.40 & 5.34 & 0.50   \\
      forward \& backward  &   {1.33} & {1.97} & {0.67} & 
    {4.29} & {5.25} & {0.64}   \\

    \hline
\end{tabular}
\end{table}

\begin{table}[t]
\centering
\setlength{\tabcolsep}{1pt}
\caption{ 
Ablation study on the intra-testing datasets {P} and {U}, \textcolor{black}{ using 2-second testing clips and different combinations of linear blocks for $G$. }
} 
\label{tab:ablation_blocks}
\footnotesize
\centering
\begin{tabular}{ >{\centering\arraybackslash}c 
                | >{\centering\arraybackslash}c c c
                | >{\centering\arraybackslash}c c c }
    \hline
    {Blocks} & 
    \multicolumn{3}{c|}{{P}} & 
    \multicolumn{3}{c}{{U}} \\
    \cline{2-7}
    (generated features)  &
    MAE$\downarrow$ & RMSE$\downarrow$ & 
    R$\uparrow$ & 
    MAE$\downarrow$ & RMSE$\downarrow$ & 
    R$\uparrow$ 
    \\\hline
    \begin{tabular}{c}$B_1$-$B_4$ ($\bb_{4}$,$\bb_{10}$)\end{tabular} 
    & 2.60 & 3.01 & 0.33 & 5.11 & 5.95 & 0.32 \\
    \midrule
    \begin{tabular}{c}$B_1$-$B_2$-$B_4$ ($\bb_{4}$,$\bb_{6}$,$\bb_{10}$)\end{tabular} 
    & 2.07 & 2.39 & 0.33 & 4.97 & 5.76 & 0.39 \\
    \midrule
    \begin{tabular}{c}$B_1$-$B_3$-$B_4$ ($\bb_{4}$,$\bb_{8}$,$\bb_{10}$)\end{tabular} 
    & 2.01 & 2.35 & 0.48 & 4.94 & 5.44 & 0.45 \\
    \midrule
    \begin{tabular}{c}$B_1$-$B_2$-$B_3$-$B_4$ ($\bb_{4}$,$\bb_{6}$,$\bb_{8}$,$\bb_{10}$)\end{tabular} 
    & {1.33} & {1.97} & {0.67} & {4.29} & {5.25} & {0.64} \\
    
    \hline
\end{tabular}
    

\end{table}

\subsubsection{On different architectures of generator} 

In Table~\ref{tab:ablation_blocks}, we explore the impact of the generator for reconstructing signals of different durations. 
In particular, we use different linear blocks to construct various progressive generators, all constrained by the same loss $\mathcal{L}_{mps\text{-}g}$. 
Note that, temporal block features $\bb_t$ produced by different linear blocks are used to generate corresponding latent rPPG features $\bg_t$ for reconstructing rPPG signals $\bar{{\mathbf{s}}}_{t}^{g}$ with different durations $t$, as shown in Figure~ \ref{fig:generator}.
The results show that when only $B_1$-$B_4$ is included, without intermediate-length signal reconstruction, the feature generator fails to extend short 2s rPPG signals to longer 10s signals of the same frequency. In contrast, reconstructing intermediate-length signals effectively facilitates accurate HR measurement.

\begin{figure*}  
    \centering
    \begin{tabular}{c c}
        \begin{minipage}{0.75\textwidth}
            \includegraphics[width=\linewidth]{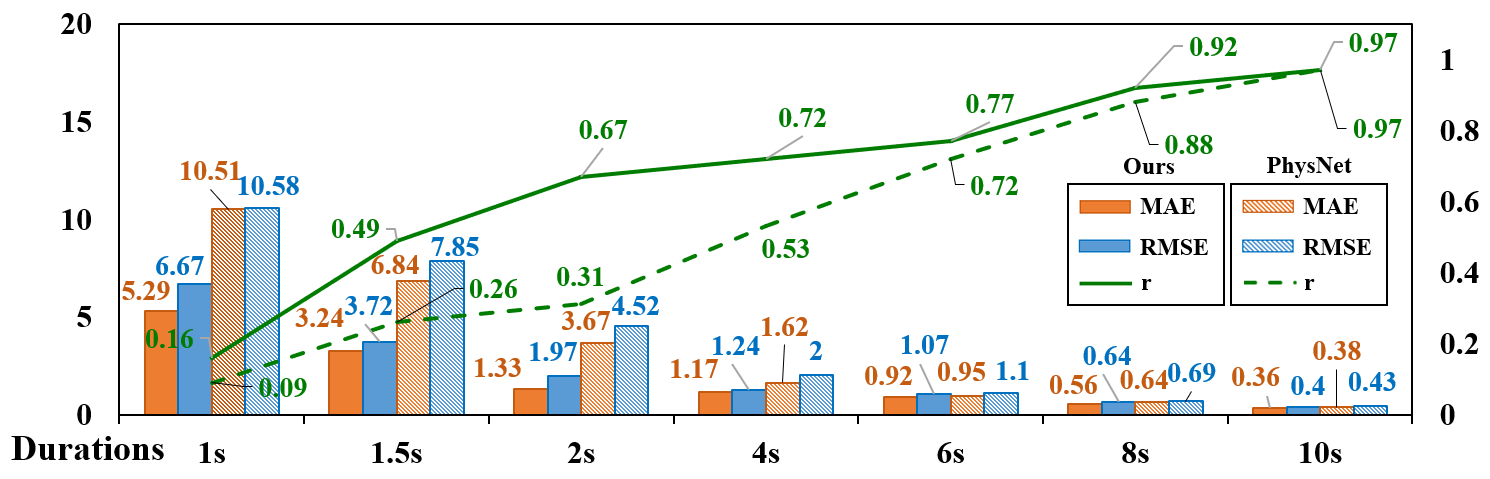}
        \end{minipage} \\ (a)
        \\ 
        \begin{minipage}{0.75\textwidth}
            \includegraphics[width=\linewidth]{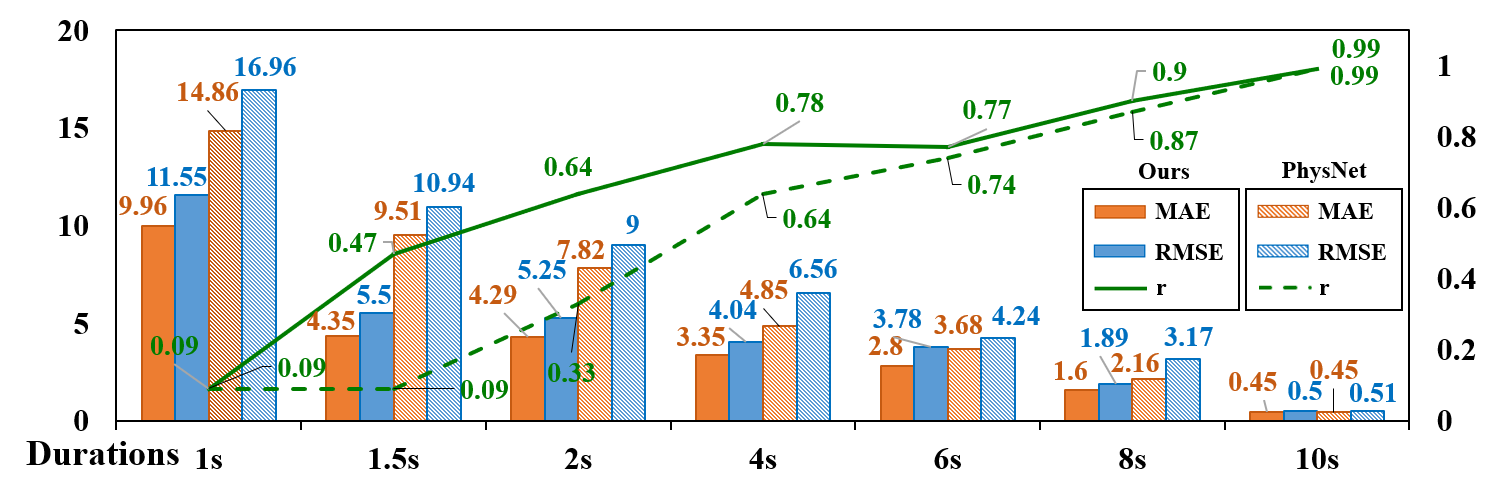}
        \end{minipage} \\ (b)
        \\ 
    \end{tabular}
\caption{   
 Impacts of testing clips with different durations on different datasets, including (a) {PURE} and (b) {UBFC-rPPG}. 
}  
\label{fig:different_video_clips}  
\end{figure*}

\begin{table}[t]
\centering
\setlength{\tabcolsep}{1pt}
\caption{
{
Ablation study for the transferability of generator $G$ on the intra-testing datasets {P} and {U}, \textcolor{black}{ using 2-second testing clips and different methods.}
}
} 
\label{tab:transferability}
\setlength{\tabcolsep}{0.2pt} 
\footnotesize
\centering
\begin{tabular}{ >{\centering\arraybackslash}c 
                | >{\centering\arraybackslash}c c c
                | >{\centering\arraybackslash}c c c
                }
    \hline
    \multirow{2}{*}
    {{Methods}} & 
    \multicolumn{3}{c|}{{P}
    } & 
    \multicolumn{3}{c }{{U}} 
    \\
    \cline{2-7}
    &
    MAE$\downarrow$ & RMSE$\downarrow$ & 
    R$\uparrow$ & 
    MAE$\downarrow$ & RMSE$\downarrow$ & 
    R$\uparrow$ 
    \\\hline
     Contrast-Phys \cite{sun2022contrast} (\textit{ECCV 22}) &  12.65 & 13.03 & 0.41 & 15.73 & 16.31 & 0.36 \\
     Contrast-Phys* \cite{sun2022contrast} (\textit{ECCV 22}) &  {9.00} & {9.71} & {0.50} & {12.27} & {12.83} & {0.41} \\
     Dual-bridging \cite{du2023dual} (\textit{CVPR 23})  & 5.82 & 6.74 & 0.42 & 7.51 & 8.67 & 0.42 \\
     Dual-bridging* \cite{du2023dual} (\textit{CVPR 23}) & {4.08} & {4.58} & {0.52} & {5.60} & {6.79} & {0.44}  \\


    \hline
\end{tabular}
\end{table}

\subsubsection{On robust transferability}
In Table~\ref{tab:transferability}, we explore the transferability of the proposed periodiciy-guided signal reconstruction. 
In particular, we consider adopting the ultra-short rPPG signals estimated by previous methods, and then include the generator to reconstruct long rPPG signals  (marked by *) for addressing the imprecise HR measurement, due to the spectral leakage issue. 
The improved results show that the proposed periodiciy-guided signal reconstruction indeed exhibits robust transferability across different remote HR measurement methods.

\subsubsection{On testing clips with different durations}

In Figure~\ref{fig:different_video_clips}, we evaluate performance on testing clips of varying durations using PhysNet \cite{yu2019physnet} as the baseline during the inference stage.
First, because 1-second clips do not exhibit one complete heartbeat cycle, we observe that performance significantly degrades when $t=1$.
Next, because 1.5-second clips exhibit at least one complete heartbeat cycle, we see that the proposed method significantly improves performance, even in the most challenging clips when $t=1.5$. 
The results in Figure~\ref{fig:different_video_clips} again confirm that performance improves with longer video durations during inference stage.

\subsubsection{On different heart rate calculation methods}
 
In Table~\ref{tab:IBI}, we compare using different HR calculation methods to calculate the corresponding HRs from the same 2-second rPPG signals $ \bar{\mathbf{s}}_2$ and estimated by the proposed method and the same 10-second rPPG signals $\bar{{\mathbf{s}}}_{t_{10}}^{g}$ reconstructed by the proposed method. 
First, in the cases of the estimated 2-second rPPG signals $\bar{\mathbf{s}}_2$,  we see that merely calculating the inter-beat intervals (IBI) \cite{yang2020classification} is insufficient for accurately detecting HR, as the 2-second rPPG signals provide only one or two intervals. 
At the same time, small variations in these intervals can lead to significant errors in HR calculation through IBI-based method.
Next, in the cases of the reconstructed 10-second rPPG signals $\bar{{\mathbf{s}}}_{t_{10}}^{g}$, since $\bar{{\mathbf{s}}}_{t_{10}}^{g}$ provides a more stable periodicity compared to $\bar{\mathbf{s}}_2$,  we observe that IBI indeed benefits from the stable periodicity of the reconstructed 10-second rPPG signals to reduce calculation errors and performs better than the estimated 2-second rPPG signals.
Finally, by comparing the PSD-based and IBI-based methods, we observe that analyzing the frequency information for HR detection is an effective approach for HR calculation.

\begin{table}[t]
\centering
\setlength{\tabcolsep}{1pt}
\caption{  
Ablation study on the intra-testing {P} and {U}, using different HR calculation methods for the same estimated rPPG signals.} 
\label{tab:IBI}
\footnotesize
\centering
\begin{tabular}{ >{\centering\arraybackslash}c 
                | >{\centering\arraybackslash}c 
                | >{\centering\arraybackslash}c c c
                | >{\centering\arraybackslash}c c c
                }
    \hline
    \multirow{2}{*}{{Testing signals}} &
    \multirow{1}{*}{{HR}} & 
    \multicolumn{3}{c|}{{P}
    } & 
    \multicolumn{3}{c }{{U}} 
    \\
    \cline{3-8}
    & {calculation} 
    &
    MAE$\downarrow$ & RMSE$\downarrow$ & 
    R$\uparrow$ & 
    MAE$\downarrow$ & RMSE$\downarrow$ & 
    R$\uparrow$ 
    \\\hline
     \multirow{2}{*}{$\bar{\mathbf{s}}_2$} & IBI-based&  3.50 & 3.86 & 0.21 & 6.50 & 8.37 & 0.36 \\
        & PSD-based  &  2.69 & 3.14 & 0.43 & {5.17} & {6.38} & {0.50}
       \\
       \hline
       \multirow{2}{*}{$\bar{{\mathbf{s}}}_{t_{10}}^{g}$} & IBI-based  &   1.42 & 2.16 & 0.61 & 4.46 & 5.42 & 0.45   \\ 
       & PSD-based  &   {1.33} & {1.97} & {0.67} & {4.29} & 
       {5.25} & {0.64}   \\

    \hline
\end{tabular}
\end{table}

\begin{table}[t]
\centering
\caption{  
Ablation study on the computational efficiency.
} 
\label{tab:speed} 
\footnotesize
\begin{tabular}{c|c|c|c}
\hline
{{Methods}}       & Params(M) & FLOPs(G) & FPS    \\ \hline
PhysFormer  \cite{yu2022physformer}   & 7.38      & 75.91    & 255.47 \\
Contrast-Phys  \cite{sun2022contrast} & {0.86}      & {51.13}    & {653.11} \\ 
\hline
{Ours}           & 1.26      & 51.30    & 629.43 \\ \hline
\end{tabular}
\end{table}

\subsubsection{On the computational efficiency}

In Table~\ref{tab:speed}, we explore the  computational efficiency of the proposed method. 
First, we observe that PhysFormer \cite{yu2022physformer}, which utilizes vision transformers as the backbone, results in a lower FPS.
Next,  since the proposed method adopts the same architecture as Contrast-Phys \cite{sun2022contrast}, using a 3DCNN as the rPPG model $T$ to estimate rPPG signals while integrating an additional generator to generate rPPG features, it maintains a relatively fast inference speed compared to Contrast-Phys \cite{sun2022contrast}.
The fast inference speed of the proposed method highlights its feasibility for real-world applications that require  quick and accurate HR estimation.

\begin{figure}[t]  
    \centering
    \begin{tabular}{c c}
        \begin{minipage}{0.5\columnwidth}
            \includegraphics[width=\linewidth]{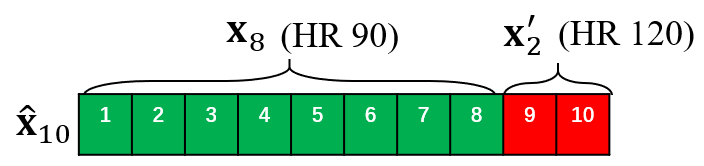}
        \end{minipage}  & 
         \begin{minipage}{0.14\columnwidth}
            \includegraphics[width=\linewidth]{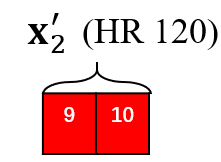}
        \end{minipage}
        \\ (a) & (b) 
    \end{tabular}
   
\caption{   
Examples of rPPG estimation under sudden HR changes using (a) 10-second and (b) 2-second sliding windows.
}
\label{fig:sliding-windows}  
\end{figure}

\begin{table}[t]
\footnotesize
\centering 
\caption{  
 HR measurement with sudden HR changes.
} 
\begin{tabular}{c|c|c|c|c}
\hline
{{Methods}} & Videos & MAE & RMSE & R   \\ \hline
RErPPG-Net \cite{hsieh2022augmentation} & \multirow{2}{*}{$\hat{\mathbf{x}}_{10}$} &  17.96   &  18.51    & 0.35 \\ \cline{1-1} \cline{3-5} 
PhysFormer \cite{yu2022physformer} & & 19.18 & 19.24 & 0.43 \\ \hline
{Ours}  & $ {\mathbf{x}}'_{2}$ & {1.95} & { 2.34} & {0.55}   \\ \hline 
\end{tabular} 

 \label{tab:sliding-windows} 
\end{table}

\begin{table}[t]
\footnotesize
\centering
\caption{  
 HR measurement  without sudden HR changes.
} 
\begin{tabular}{c|c|c|c|c}
\hline
{{Methods}} & Videos & MAE & RMSE & R \\ \hline  
RErPPG-Net \cite{hsieh2022augmentation} & \multirow{3}{*}{$ {\mathbf{x}}_{10}$}  & 0.38 & 0.54 & 0.96 \\ \cline{1-1} \cline{3-5}
PhysFormer \cite{yu2022physformer} & & 0.40 & 0.44 & 0.96  \\ \cline{1-1} \cline{3-5} 
{Ours}  & & {0.36} & {0.40} & {0.97}  \\ \hline 
{{Methods}} & Videos & MAE & RMSE & R  \\ \hline  
RErPPG-Net \cite{hsieh2022augmentation} & \multirow{3}{*}{$ {\mathbf{x}}_{2}$} & 6.25 & 7.83 & 0.47 \\ \cline{1-1} \cline{3-5}
PhysFormer \cite{yu2022physformer}  & & 6.56 & 8.29 & 0.43   \\ \cline{1-1} \cline{3-5} 
{Ours}  & & {1.33} & {1.97} & {0.67}  \\ \hline 
\end{tabular}
\label{tab:10s-windows} 
\end{table}

\subsubsection{On simulating sudden heart rate changes}

\textcolor{black}{As shown in Table~\ref{tab:variations-HR}, existing public rPPG datasets are generally collected under relatively stable physiological conditions, where the heart rate remains nearly constant throughout each video.
To the best of our knowledge, there is currently no publicly available rPPG dataset that contains substantial heart-rate variations within the same video sequence.}
To validate the significance and advantages of our  method, we conduct a supplementary experiment simulating a sudden heart rate (HR) change. Following the sampling strategy in [5], we augment each 10s facial video $\mathbf{x}_{10}$ from dataset {P} by increasing its HR by 33\% to obtain $\mathbf{x}'_{10}$. We then construct a new 10s video $\hat{\mathbf{x}}_{10}$ by concatenating the first 8s clip $\mathbf{x}_{8}$ of $\mathbf{x}_{10}$ with the final 2s clip $\mathbf{x}'_{2}$ of $\mathbf{x}'_{10}$ to simulate a sudden HR change, as shown in Figure~\ref{fig:sliding-windows} (a).
Table~\ref{tab:sliding-windows} presents the sudden HR change measured by applying   existing rPPG methods [8,22] to both $\hat{\mathbf{x}}_{10}$ and $\mathbf{x}'_{2}$, where the ground truth for this comparison is the modified HR  in $\mathbf{x}'_{2}$ (Figure~\ref{fig:sliding-windows} (b)).
Notably, the application of a continuous 10s sliding window ($\hat{\mathbf{x}}_{10}$) for estimating ultra-short rPPG information results in measured HR values primarily dominated by the first 8s of the video (highlighted in green), failing to promptly and accurately capture the actual HR change within the ultra-short 2s segment (highlighted in red), as shown in Figure~\ref{fig:sliding-windows}.
\textcolor{black}{
In contrast, since our method directly analyzes the ultra-short 2-second clip (${\mathbf{x}}'_{2}$) containing sudden HR changes, our method is able to promptly capture the actual heart-rate variation occurring within the clip, thereby enabling accurate and timely HR estimation.}


In addition, Table~\ref{tab:10s-windows} presents the performance of  existing rPPG methods [8,22] in measuring HRs from both ${\mathbf{x}}_{10}$ and ${\mathbf{x}}_{2}$ without sudden HR changes.
While these methods demonstrate promising results with 10s clips ${\mathbf{x}}_{10}$, their performance significantly degrades when evaluated on 2s  clips ${\mathbf{x}}_{2}$.
In contrast, our method maintains promising performance even with the shorter 2s  clips ${\mathbf{x}}_{2}$.

\subsubsection{Visualization} 
\begin{figure}[t]  
    \centering
    \begin{tabular}{cc} 
    \fbox{\includegraphics[width=0.36\columnwidth]{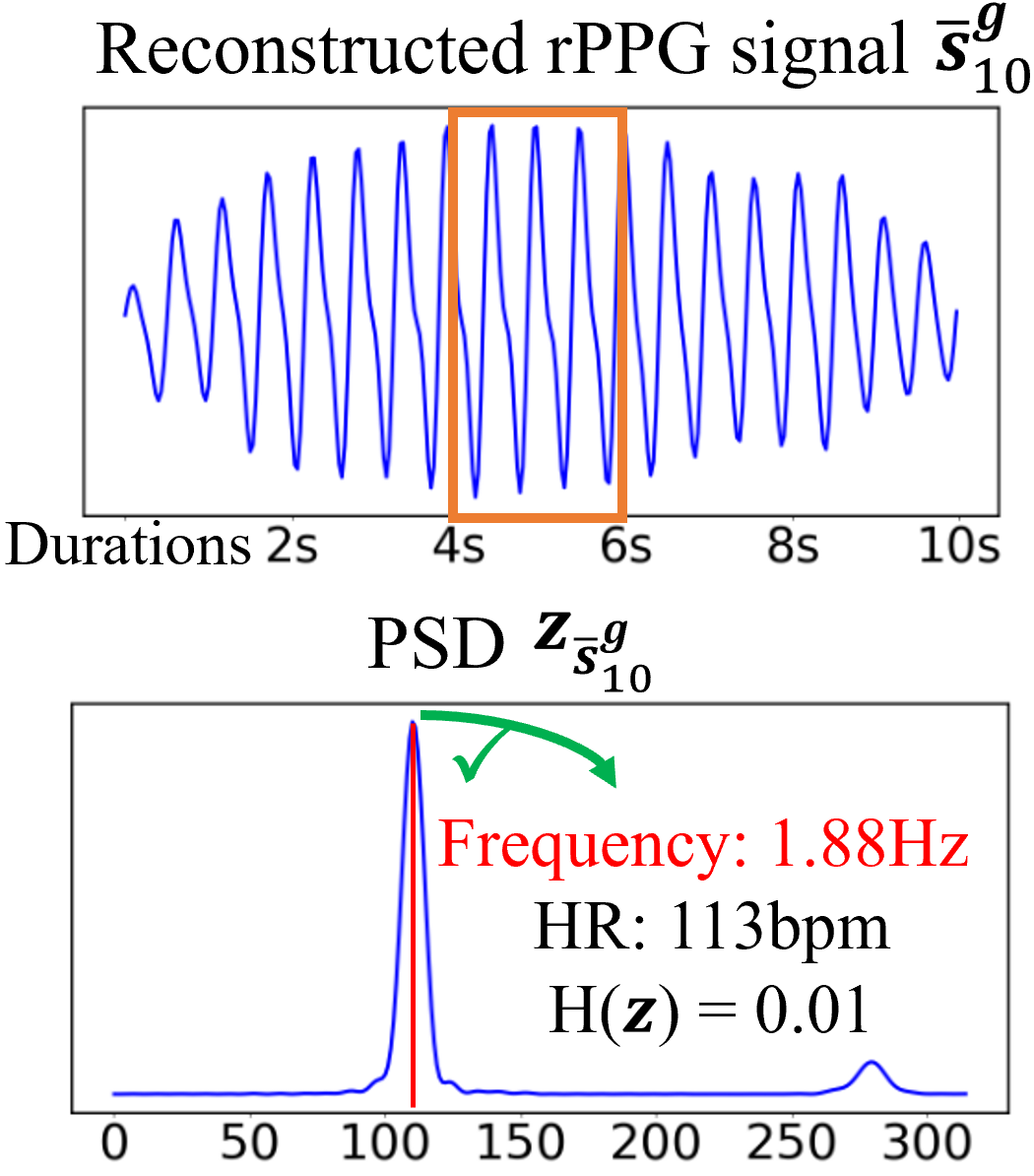}} &
    \fbox{\includegraphics[width=0.36\columnwidth]{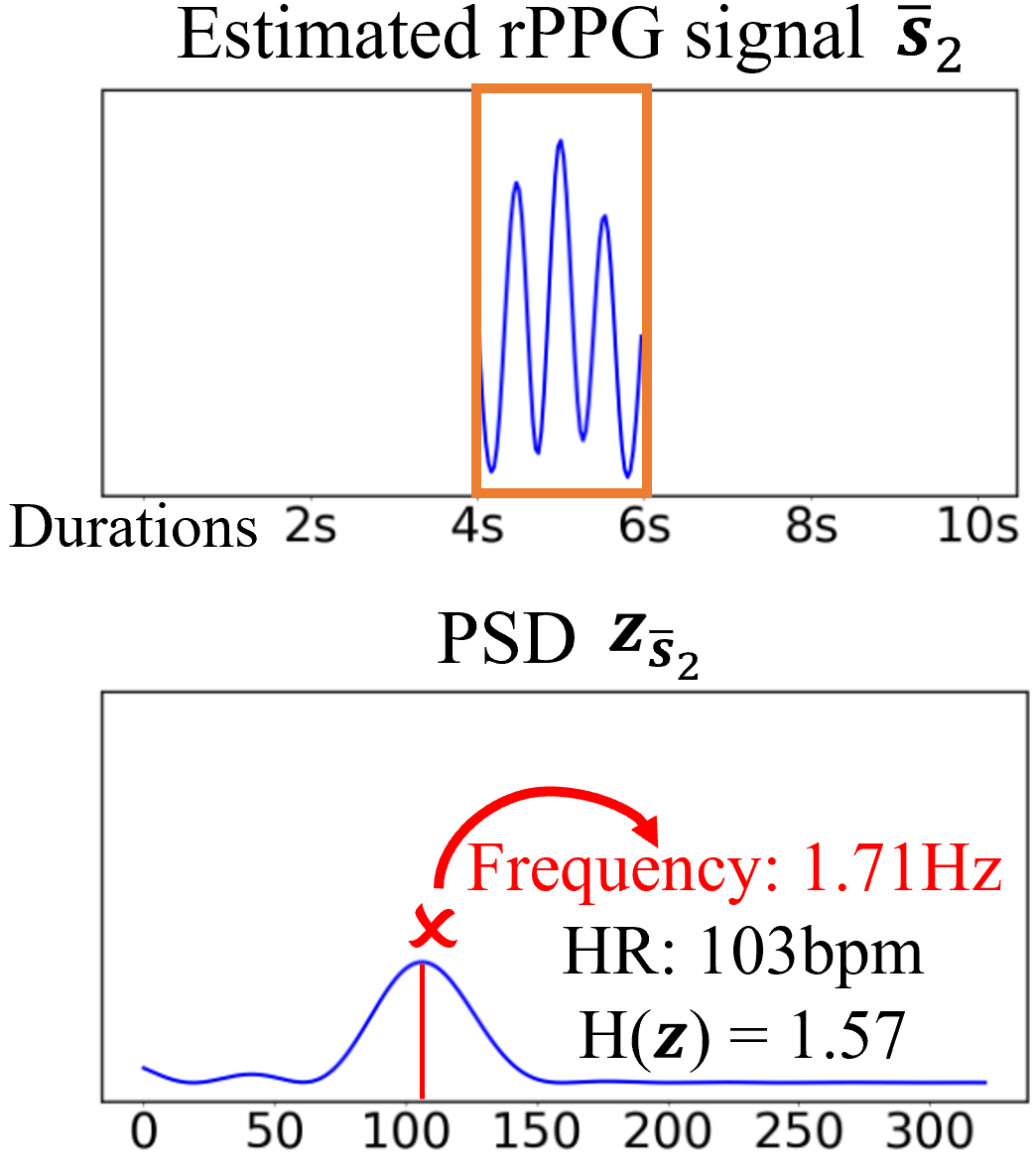}} 
    \\ (a)&(b)  
    \end{tabular} 
   
\caption{   
\textcolor{black}{Examples of rPPG signals and their corresponding PSDs: } (a) the reconstructed 10-second rPPG signal and (b) the estimated 2-second rPPG signal.
}
\label{fig:issue}  
\end{figure}

In Figure~\ref{fig:issue}, we visualize the reconstructed rPPG signal to demonstrate the efficacy of the proposed method.
Note that Figures \ref{fig:issue_in_GT} and ~\ref{fig:issue} show the ground truth rPPG signals and the reconstructed/estimated rPPG signals from the same sample, respectively.
First, thanks to the proposed weighted spectral-based cross-entropy loss, we observe that the estimated 2-second and reconstructed 10-second rPPG signals exhibit lower entropy compared to the ground truth 2-second and 10-second rPPG signals.
This indicates that the energy of the estimated/reconstructed rPPG signals is more concentrated at specific frequencies, enabling more accurate HR measurement \cite{speth2023non}.
Next, due to the spectral leakage issue, we see that the estimated 2-second rPPG signal exhibits imprecise PSDs, as shown in Figure~\ref{fig:issue} (b). 
Furthermore, as observed in Figure~\ref{fig:issue} (a), the reconstructed rPPG signal exhibits the same periodicity as the ground truth rPPG signal in Figure~\ref{fig:issue_in_GT} (a)  and effectively  addresses the spectral leakage issue to produce precise PSDs and accurate HR measurement.

\begin{table}[t]
\centering
\setlength{\tabcolsep}{1pt}
\caption{ 
\textcolor{black}{Generalization evaluation on {U} and {P}, using 10-second testing clips.}
} 
\label{tab:experiments-10s} 
\footnotesize
\centering
  \color{black}
\begin{tabular}{ >{\centering\arraybackslash}c 
                | >{\centering\arraybackslash}c c c
                | >{\centering\arraybackslash}c c c
                }
    \hline
    \multirow{2}{*}
    {{Methods}}  & 
    \multicolumn{3}{c|}{{P}
    } & 
    \multicolumn{3}{c }{{U}} 
    \\
  
    \cline{2-7}
    { }
      &
    MAE$\downarrow$ & RMSE$\downarrow$ & 
    R$\uparrow$ & 
    MAE$\downarrow$ & RMSE$\downarrow$ & 
    R$\uparrow$ 
    \\\hline
    Contrast-Phys \cite{sun2022contrast}  &  1.00 & 1.40 & 0.99 & 0.64 & 1.00 & 0.99
        \\
    RErPPG-Net \cite{hsieh2022augmentation}  &  0.71 & 1.48 & 0.96 & 0.71 & 1.48 & 0.96
        \\
    PhysFormer \cite{yu2022physformer}  &  1.10 & 1.75 & 0.99 & 0.40 & 0.71 & 0.99
        \\
    EfficientPhys \cite{liu2023efficientphys}  &  1.33 & 5.99 & 0.97 & 1.14 & 1.81 & 0.99
        \\
    RhythmFormer \cite{zou2025rhythmformer}  &  3.14 & 9.65 & 0.92 & 1.60 & 3.37 & 0.98
        \\
       \hline
    Ours &  \textbf{0.36} & \textbf{0.40} & \textbf{0.97} & 
    \textbf{0.45} & \textbf{0.50} & \textbf{0.99}   \\
    \hline
\end{tabular}
\end{table}

\subsubsection{\textcolor{black}{Evaluation on long testing clips}}

\textcolor{black}{ 
In Table~\ref{tab:experiments-10s}, we conduct additional experiments using 10-second testing clips to further evaluate whether the proposed method remains effective under longer video durations, where the physiological information within the same video is relatively stable, as evidenced by the heart-rate variation statistics reported in Table~\ref{tab:variations-HR}. This evaluation setting is also consistent with existing public rPPG benchmarks, where substantial heart-rate variations within a single video sequence are rarely observed. The strong performance achieved on 10-second testing clips further demonstrates that the proposed method not only benefits ultra-short rPPG estimation but also generalizes effectively to longer video sequences with relatively stable physiological information.
}

\subsection{\textcolor{black}{Limitations and future improvement}}

\begin{figure}[t]  
    \centering
    \includegraphics[width=9cm]{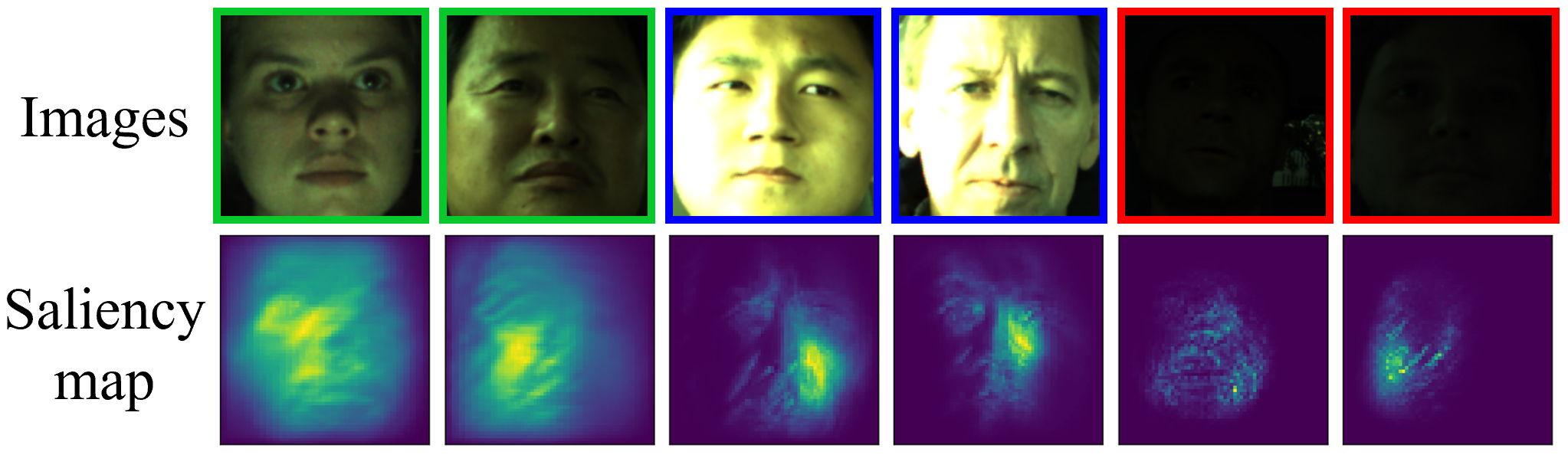} 
\caption{   
\textcolor{black}{
Samples from \textbf{MRNIRP Car} \cite{nowara2020near} under varying illumination and their corresponding saliency maps. Green, red, and blue boxes denote normal, low-light, and bright illumination, respectively. A well-trained estimator should produce saliency maps with strong responses concentrated on the face.
}} 
\label{fig:limitation}
\vspace{-0.4cm}
\end{figure}

\textcolor{black}{In Figure~\ref{fig:limitation},  we further evaluate the proposed method under challenging illumination conditions, including bright and low-light environments.
Since rPPG estimation relies on subtle skin color variations induced by blood volume changes, its performance is inherently sensitive to illumination variations.
Although the proposed method can effectively identify facial regions containing rPPG-related physiological information under normal illumination, the corresponding saliency maps become less concentrated and exhibit weaker responses under extreme lighting conditions, particularly in low-light scenarios.
This observation suggests that severe illumination variations may degrade the quality of the extracted physiological features and reduce the model's ability to consistently focus on informative facial regions. In future work, we will focus on improving illumination robustness through illumination-invariant feature learning and adaptive enhancement strategies, thereby enabling more reliable rPPG estimation in varying real-world environments.
}

\section{Conclusion}
In this paper, we proposed a novel periodicity-guided method to address two key challenges in heart rate (HR) measurement from ultra-short video clips: insufficient heartbeat cycles in ultra-short clips and HR estimation inaccuracy due to spectral leakage. 
First, we proposed an effective periodicity-guided rPPG estimation to accurately estimate rPPG signals from ultra-short video clips by enforcing consistent periodicity between  rPPG signals estimated from different durations of the same video.  
Next, to address the spectral leakage issue, we proposed a novel periodicity-guided signal reconstruction by incorporating a generator to reconstruct longer rPPG signals from ultra-short ones while preserving their periodic consistency to enable more precise PSD estimation and accurate HR measurement in the rPPG model.  
Extensive experiments demonstrated that the proposed method effectively measures HR from ultra-short clips and outperforms previous rPPG estimation methods to achieve state-of-the-art performance. 
\textcolor{black}{Future work will focus on improving illumination robustness through illumination-invariant feature learning and adaptive enhancement strategies to enable more reliable rPPG estimation in diverse real-world environments.}

\section{Acknowledgment}

This work was supported in part by the National Science and Technology Council grant 112-2221-E-007-082-MY3 of Taiwan and the Natural Science Foundation of Fujian Province, China under   Grant 2026J008139. 

\bibliographystyle{elsarticle-num-names} 
\bibliography{egbib}





\end{document}